\documentclass[lettersize,journal]{IEEEtran}

\usepackage{amsmath,amsfonts}
\usepackage{algpseudocode,algorithm}
\usepackage{array}
\usepackage{textcomp}
\usepackage{stfloats}
\usepackage{url}
\usepackage{verbatim}
\usepackage{graphicx}
\usepackage{cite}
\usepackage{multirow}

\usepackage{siunitx}
\usepackage{balance}
\ifCLASSOPTIONcompsoc
\usepackage[caption=false, font=normalsize, labelfont=sf, textfont=sf]{subfig}
\else
\usepackage[caption=false, font=footnotesize]{subfig}
\fi
\graphicspath{{./fig/}}

\usepackage{blindtext}
\usepackage{hyperref}

\begin{document}
	
\title{
MPPI-IPDDP: A Hybrid Method of Collision-Free Smooth Trajectory Generation for Autonomous Robots
}
	
\author{
{Min-Gyeom Kim}, {Minchan Jung}, {JunGee Hong} and {Kwang-Ki~K.~Kim}
\thanks{The authors are with the Department of Electrical and Computer Engineering, Inha University, Incheon 22212, Republic of Korea.}
\thanks{Corresponding author: K.-K.~K. Kim ({\tt kwangki.kim@inha.ac.kr})}
\thanks{This research was supported by the Basic Science Research Program through the National Research Foundation of Korea (NRF) funded by the Ministry of Education (NRF- 2020R1F1A1076404, NRF-2022R1F1A1076260).}
}
	
	
	
\maketitle


\begin{abstract}
This paper presents a hybrid trajectory optimization method designed to generate collision-free, smooth trajectories for autonomous mobile robots. By combining sampling-based Model Predictive Path Integral (MPPI) control with gradient-based Interior-Point Differential Dynamic Programming (IPDDP), we leverage their respective strengths in exploration and smoothing. The proposed method, MPPI-IPDDP, involves three steps: First, MPPI control is used to generate a coarse trajectory. Second, a collision-free convex corridor is constructed. Third, IPDDP is applied to smooth the coarse trajectory, utilizing the collision-free corridor from the second step. To demonstrate the effectiveness of our approach, we apply the proposed algorithm to trajectory optimization for differential-drive wheeled mobile robots and point-mass quadrotors. In comparisons with other MPPI variants and continuous optimization-based solvers, our method shows superior performance in terms of computational robustness and trajectory smoothness.
\\[2mm] 
\indent{\tt Code:} \href{https://github.com/i-ASL/mppi-ipddp}{\tt https://github.com/i-ASL/mppi-ipddp} \\[1mm]
\indent{\tt video:} \href{https://youtu.be/-oUAt5sd9Bk}{\tt https://youtu.be/-oUAt5sd9Bk}
\end{abstract}
	
	\begin{IEEEkeywords}
		Trajectory optimization,
		Planning, 
		Obstacle avoidance,
		Variational inference,
		Model predictive path integral,
		Differential dynamic programming,
		Collision-free corridors.
	\end{IEEEkeywords}

	
\section{Introduction}\label{sec:intro}
Path planning is a critical problem for autonomous vehicles and robots. Several considerations need to be addressed simultaneously in robot path planning and navigation, such as specifying mission goals, ensuring dynamic feasibility, avoiding collisions, and considering internal constraints.

Optimization-based methods for path planning can explicitly handle these tasks. Two popular optimal path planning methods for autonomous robots are gradient-based and sampling-based methods. Gradient-based methods assume that the objective and constraint functions in the planning problem are differentiable, allowing for a fast, locally optimal smooth trajectory. These methods typically rely on nonlinear programming solvers such as IPOPT~\cite{wachter2006implementation} and SNOPT~\cite{gill2005snopt}.
On the other hand, sampling-based methods do not require function differentiability, making them more suitable for modeling obstacles of various shapes. Additionally, they naturally perform exploration, helping escape local optima. However, derivative-free sampling-based methods often result in coarse (e.g., zigzag) trajectories. For example, RRT-based methods can generate coarse trajectories~\cite{kuffner2000rrt}. To balance the pros and cons of both methods, a hybrid approach combining them, as proposed in~\cite{campos2017hybrid}, can be considered.

The optimization-based trajectory generation architecture known as model predictive control (MPC) has been extensively applied to robotic trajectory generation and planning problems~\cite{katrakazas2015real,paden2016survey}. 
Deep reinforcement learning-based trajectory generation for mobile robots is another popular approach~\cite{chai2022design}.
A comparison of the continuous optimal control and reinforcement learning frameworks for trajectory generation of autonomous drone racing is provided in~\cite{song2023reaching}. Combining MPC with learning schemes has drawn noticeable attention to the robotics and control community~\cite{amos2018differentiable,romero2024actor,reiter2024ac4mpc}. Using the property of differential flatness, a robotic trajectory optimization problem can be converted to finite-dimensional parametric optimization~\cite{milam2003real}. 
	
\begin{figure}[tbp]
		\centerline{\includegraphics[width=.9\linewidth]{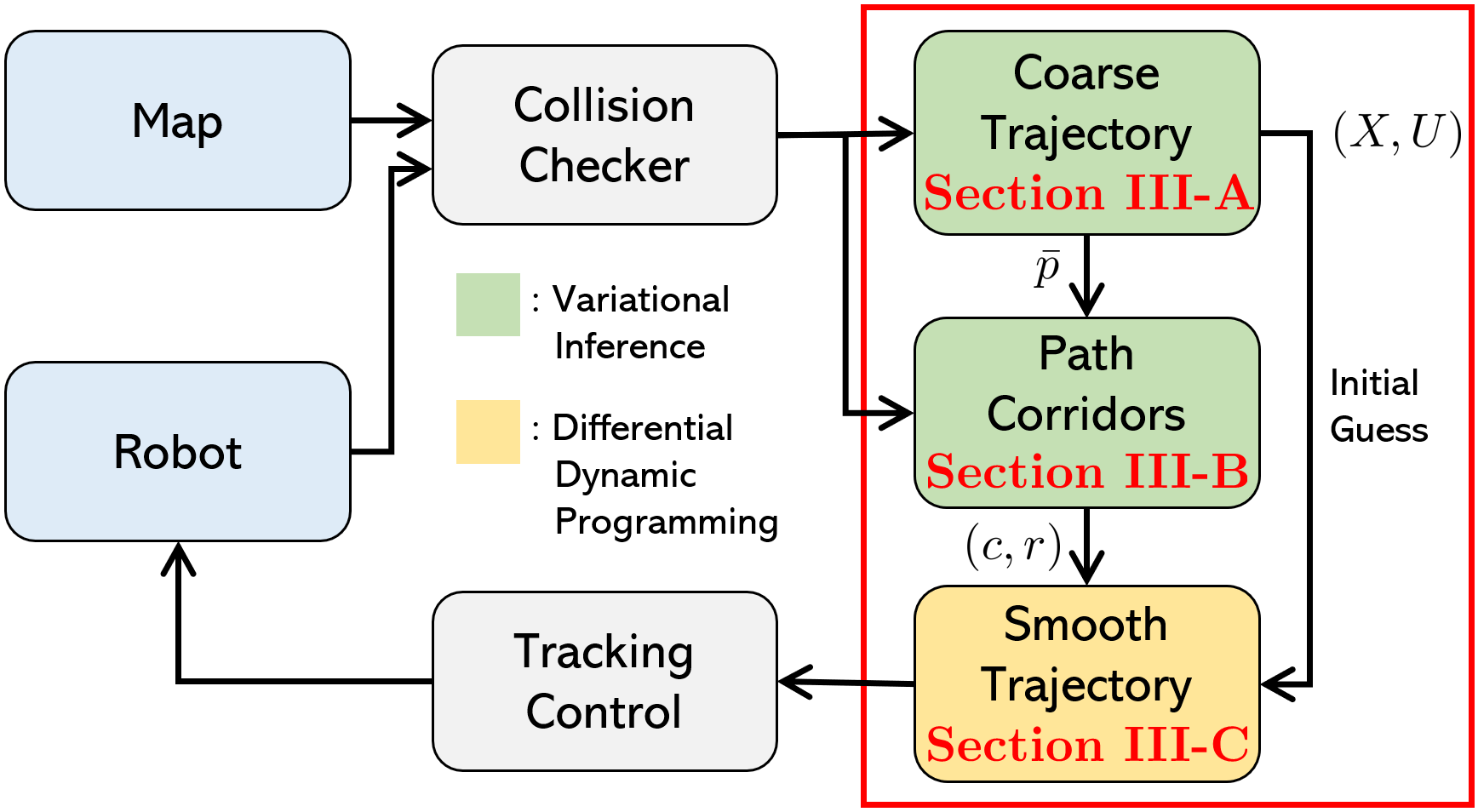}}\vspace{-2.5mm}
		\caption{A method for collision-free smooth path planning. The contents in the red box are subjects in this paper.}
		\label{fig:chart}\vspace{-2mm}
	\end{figure}

This paper proposes a hybrid trajectory optimization method that modularly incorporates sampling-based and gradient-based methods. Fig.~\ref{fig:chart} illustrates the structure of the proposed collision-free smooth path planning approach. Our method generates a coarse trajectory and path corridors using sampling-based optimization via variational inference (VI). Subsequently, a smooth trajectory is obtained through gradient-based optimization via the differential dynamic programming (DDP) scheme. We assume that a collision checker is available to determine whether a collision has occurred.
	
Variational inference (VI) refers to a class of optimization-based approaches for approximating posterior distributions, making Bayesian inference computationally efficient and scalable~\cite{zhang2018advances,murphy2022probabilistic}. The recently proposed model predictive path integral (MPPI) is a sampling-based planning method that uses the VI framework~\cite{williams2018information,wang2021variational}. In essence, MPPI samples random trajectories around a nominal trajectory, assigns weights based on cost, and updates the nominal trajectory using the weighted average. In this paper, MPPI is used to generate a coarse trajectory for exploration while avoiding collisions.

While methods such as RRT and dynamic programming (DP) can achieve collision-free rough trajectory planning, we select MPPI control due to its suitability for real-time trajectory generation as a local planner, whereas RRT-like methods are often used as global planners. MPPI offers significant computational efficiency, allowing it to operate in real-time, which is critical for continuous control tasks. Additionally, MPPI inherently incorporates system dynamics within its rollout-based framework, providing a more seamless integration between trajectory planning and control. In contrast, RRT-like methods, while effective for finding rough trajectories, suffer from unpredictable computation times, which pose challenges for real-time controller design. This makes MPPI a better fit for our goal of real-time, dynamically feasible trajectory generation.
	
To smooth the coarse trajectory with gradient-based optimization, we introduce the concept of path corridors, a popular scheme in the literature~\cite{geraerts2007corridor,campos2017hybrid,schafer2021computation}. Path corridors are collections of convex collision-free regions guiding a robot toward a goal position. Unlike previous works, we use simple sampling-based VI framework to construct these corridors.
	
To achieve a smooth trajectory, we apply the differential dynamic programming (DDP) framework for gradient-based optimization. DDP-based approaches, including the iterative linear quadratic regulator (iLQR), have become popular for nonlinear optimal control problems and have been applied in many contexts of planning and nonlinear model predictive control for autonomous systems~\cite{chai2020overview,aoyama2021constrained}. DDP relies on Bellman's principle of optimality and the necessary conditions for optimal control problems, assuming all functions defined in the problem are smooth or at least twice continuously differentiable.

Since original DDP approaches do not consider system state and input constraints, various methods have been developed to handle constraints efficiently in DDP. The augmented Lagrangian (AL) method is used in~\cite{howell2019altro}, while the Karush-Kuhn-Tucker (KKT) condition is employed in~\cite{xie2017differential}. In~\cite{aoyama2021constrained}, a method combining the AL method with the KKT condition is proposed. The interior point differential dynamic programming (IPDDP) algorithm~\cite{pavlov2021interior}, used in this work, is based on the KKT condition. IPDDP, summarized in Section~\ref{sec:preliminaries}, incorporates all Lagrangian and barrier terms into the Q-function and solves a minimax problem.

The main contributions of this paper can be summarized as follows:
\begin{itemize}
	\item
	{\em Hybrid Path Planning Method:} A novel hybrid path planning method is proposed. This method generates collision-free smooth trajectories by integrating sampling-based trajectory optimization using Model Predictive Path Integral (MPPI) and gradient-based smooth optimization (IPDDP).
	\item
	{\em Collision-Free Convex Path Corridors:} WA new method for constructing collision-free convex path corridors is introduced. This method leverages sampling-based optimization with variational inference to ensure the path is safe from obstacles.
	\item 
	{\em Effectiveness Demonstration:} MPPI-IPDDP is demonstrated to be effective through two numerical case studies. These studies showcase the practical applicability and performance of the method in generating feasible and smooth trajectories.
	\item
	{\em Open-Sourced Codes:} The C++ and MATLAB codes for the proposed MPPI-IPDDP solver are made available as open-source. This allows readers to replicate the results presented in the paper and customize the solution for their own robotic applications.
\end{itemize}

\indent
The remainder of this paper is organized as follows: Section~\ref{sec:preliminaries} reviews sampling-based optimization via variational inference and IPDDP. Section~\ref{sec:method} presents our path planning method, MPPI-IPDDP, for generating collision-free smooth trajectories. In Section~\ref{sec:case}, the effectiveness of the proposed MPPI-IPDDP is demonstrated through simulations in various environments and compared with other MPPI variants and NLP-based solvers. Section~\ref{sec:discussion} discusses the remaining challenges and practical limitations. Finally, Section~\ref{sec:conclusion} concludes the paper with suggestions for future work.
	
\section{Preliminaries}\label{sec:preliminaries}
\subsection{Sampling-based Optimization via Variational Inference}
\label{sec:preliminaries:viso}
An optimization problem can be reformulated as an inference problem and solved using the variational inference method~\cite{kappen2012optimal,levine2018reinforcement}. To achieve this, we introduce a binary random variable $o$ that indicates optimality, where $p(o=1)$ represents the probability of optimality. For simplicity, we denote this probability as $p(o)$.
	
In this paper, we consider two different cases of variational inference (VI) for stochastic optimal control: VI for finite-dimensional optimization, where the decision variable is a parameter vector, and VI for trajectory optimization, where the goal is to generate an optimal trajectory for a control system.
The baseline methodology for these VI approaches is based on Model Predictive Path Integral (MPPI) control, which serves as a sampling-based framework for stochastic control problems~\cite{williams2018information, wang2021variational, kazim2024recent}. MPPI leverages importance sampling techniques to iteratively update control policies, making it well-suited for handling the probabilistic nature of the control tasks in both finite-dimensional optimization and trajectory optimization contexts.

\subsubsection{VI for Finite-dimensional Optimization}
\label{sec:preliminaries:viso:vector}
Let $\theta$ be a vector of decision variables. For variational inference corresponding to stochastic optimization or optimal control, the goal is to find the target distribution $q^*$\footnote{We will abuse the terminology of distributions (probability measure) and probability density functions.} defined as
\begin{equation}
	q^*(\theta) = p(\theta ~\!|\!~ o) = \frac{p(o ~\!|~\!\theta)p(\theta)}{\int  p(o ~\!|~\!\theta)p(\theta)d\theta}
\end{equation}
	
Let $L(\theta) = p(o ~\!|\!~ \theta)$ be the likelihood function and $\tilde{q}^*$ be the empirical approximation of $q^*$ that is computed from samples $\{\theta_1,\dots,\theta_N\}\sim p(\theta)$ that are drawn from the prior $p(\theta)$. Then, $\tilde{q}^*$ can be represented as
\begin{equation}
	\tilde{q}^*(\theta) = \sum_{i=1}^{N} w_i\delta(\theta = \theta_i)
	\enspace\!,\quad\!\! w_i = \frac{L(\theta_i)}{\sum_{j=1}^{N}  L(\theta_j)}
\end{equation}
where $\delta$ is the Dirac delta function, and $N$ is the number of samples.
Replacing $q^*$, we approximate $\tilde{q}^*$ with the forward KL divergence:
\begin{equation}
	\begin{aligned}
	\pi^* &= \underset{\pi}{\text{arg\,min}}\enspace \mathcal{D}_{\text{KL}}(\tilde{q}^*(\theta) \| \pi(\theta))  \\
		&= \underset{\pi}{\text{arg\,max}}\enspace \mathbb{E}_{\theta\sim\tilde{q}^*(\theta)}\left[ \log\pi(\theta) \right] \,.
	\end{aligned}
\end{equation}

If a normal distribution is chosen for parameterizing the policy $\pi$, then we get the closed-form solution for the optimal policy $\pi^*=\mathcal{N}(\mu,\Sigma)$ where
\begin{equation}\label{eq:vector_vi}
	\mu = \sum_{i=1}^{N}w_i\theta_i \enspace\!,\quad\!\! \Sigma = \sum_{i=1}^{N}w_i(\theta_i - \mu)(\theta_i - \mu)^\top
\end{equation}
	
In this paper, this VI-based stochastic optimization method is used for constructing collision-free convex path-corridors in Section~\ref{sec:method:corridor}.
	
\subsubsection{VI for Trajectory Optimization}
\label{sec:preliminaries:viso:traj}
Let $\tau=(X,U)$ be a trajectory consisting of a sequence of controlled states $X=(x_0,\dots,x_T)$ and a sequence of control inputs $U=(u_0,\dots,u_{T-1})$ over a finite time-horizon $T$. The goal is to find the target distribution $q^*(\tau) = p(X ~\!|\!~ U)q^*(U)$ where $p(X ~\!|\!~ U)$ represents stochastic dynamics:
\begin{equation}\label{eq:target_traj}
	\begin{aligned}
	q^*(\tau) &\!=\! \underset{q(\tau)}{\text{arg\,min}}\enspace \!\! \mathcal{D}_{\text{KL}}(q(\tau) \| p(\tau | o)) \\
	               &\!=\! \underset{q(\tau)}{\text{arg\,min}}\enspace \!\! \mathcal{D}_{\text{KL}}(q(U) \| p(U)) \!-\! \mathbb{E}_{\tau\sim q(\tau)}\! \left[ \log p(o | \tau) \right]  .
	\end{aligned}
\end{equation}

Let $L(U) = \mathbb{E}_{X\sim p(X ~\!|\!~ U)} \! \left[ \log p(o ~\!|\!~ \tau) \right]$. Then $q^*(\tau)$ can be rewritten as
\begin{equation}
	\label{eq:vi_traj}
	\begin{aligned}
		q^*(U) = \underset{q(U)}{\text{arg\,min}} \,
		& \mathcal{D}_{\text{KL}}(q(U) \| p(U)) - \mathbb{E}_{U\sim q(U)} \! \left[ L(U) \right] \\
		\text{s.t.} \,
		& \int q(U)dU = 1 \,.
	\end{aligned}
\end{equation}
The closed-form solution for the above optimization is given by 
\begin{equation}
	q^*(U) = \frac{\exp(L(U))p(U)}{\int\exp(L(U))p(U)dU}
\end{equation}
Let $\tilde{q}^*$ be the empirical distribution of $q^*$ approximated with samples $\{U_1,\dots,U_N\}\sim p(U)$ drawn from the prior $p(U)$. Then, $\tilde{q}^*$ can be represented as
\begin{equation}
	\tilde{q}^*(U) = \sum_{i=1}^{N} w_i\delta(U = U_i)\,, \ w_i = \frac{\exp(L(U_i))}{\sum_{j=1}^{N} \exp(L(U_j))}
\end{equation}
Replacing $q$, we approximate $\tilde{q}^*$ with the forward KL divergence.
\begin{equation}
	\pi^* = \underset{\pi}{\text{arg\,min}}\enspace \mathcal{D}_{\text{KL}}(\tilde{q}^*(U) ~\!\|\!~ \pi(U))
\end{equation}

If the normal distribution is chosen for $\pi$, then we get the closed form solution of $\pi^*=\mathcal{N}(\mu,\Sigma)$ where
\begin{equation}\label{eq:traj_vi}
	\mu = \sum_{i=1}^{N}w_iU_i \enspace\!,\quad\!\!
	\Sigma = \sum_{i=1}^{N}w_i(U_i - \mu)(U_i - \mu)^\top
\end{equation}
In this paper, this VI-based trajectory optimization is applied for MPPI \cite{williams2018information,wang2021variational} to generate a locally optimal trajectory in Section~\ref{sec:method:mppi}.
	
\subsubsection{Additional Notes}
\label{sec:preliminaries:viso:notes}
One of the most common choices for the likelihood function is $p(o ~\!|\!~ \cdot) = \exp(-\gamma J(\cdot))$ where $J(\cdot)$ is a cost function and $\gamma >0$ is known as the inverse temperature. With this likelihood function, the weight $w_{i}$~in Sections~\ref{sec:preliminaries:viso:vector} and \ref{sec:preliminaries:viso:traj} can be interpreted as the likelihood ratio corresponding to the sampled candidate $\theta_{i}$ or $U_{i}$, respectively. This implies that the lower the value of $J$ the higher the likelihood of being optimal at an exponential rate.
	
Since this sampling-based optimization scheme is iterative, the distribution $\pi$ should influence the prior $p$ in the next iteration, ensuring that $\pi$ eventually reaches a locally optimal point. In this paper, we assume normal distributions for both the prior and posterior, propagating only the mean $\mu$ while using a fixed covariance $\Sigma$. We do not perform empirical adaptation as outlined in (\ref{eq:vector_vi}) and (\ref{eq:traj_vi}).
	
	
\vspace{-4mm}
\subsection{Interior Point Differential Dynamic Programming}
\label{sec:preliminaries:alddp}
IPDDP introduced in~\cite{pavlov2021interior} can be used to solve a standard discrete-time optimal control problem (OCP) given as
\begin{equation}\label{eq:ocp}
	\begin{aligned}
		\underset{u_{0:T-1}}{\text{minimize}} \quad &l_f(x_T) + \sum_{t=0}^{T-1} l_t(x_t,u_t)\\
		\text{subject to} \quad 
		& x_{t+1} = f_t(x_t,u_t),\,x_0 = x_{\rm init}\\
		& g_t(x_t,u_t) \leq 0
	\end{aligned}
\end{equation}
where the variables $x_t \in \mathbb{R}^n$ and $u_t \in \mathbb{R}^m$ are the system state and the control input vector at time-step $t$, respectively, and $x_{\rm init} $ is the initial condition for the control system. Let denote the decision vector as $U := u_{0:T-1} = [ u_{0}^\top, u_{1}^\top, \cdots, u_{T-1}^\top ]^\top \in \mathbb{R}^{nT}$ that is the concatenation of sequential control inputs over a time horizon $T$. The real-valued functions $l_f: \mathbb{R}^n \rightarrow \mathbb{R}$ and $l_t: \mathbb{R}^n \times \mathbb{R}^m \rightarrow \mathbb{R}$ are the final and stage cost functions, respectively, and $f_t: \mathbb{R}^n \times \mathbb{R}^m \rightarrow \mathbb{R}^n$ defines the controlled state transitions. The vector-valued function $g_t: \mathbb{R}^n \times \mathbb{R}^m \rightarrow \mathbb{R}^k$ defines inequality constraints where $k$ denotes the number of constraints. All functions defined in~\eqref{eq:ocp} are assumed to be twice continuously differentiable.
	
In dynamic programming perspectives, the OCP~\eqref{eq:ocp} can be converted into the Bellman equation form at time $t$ with a given state $x_t$ as follows:
\begin{equation}
	\begin{aligned}
		V_t(x_t) = \underset{u_t,s_t}{\text{min}} 
		\enspace &l_t(x_t,u_t) + V_{t+1}(f_t(x_t,u_t))\\
		\text{subject to} \enspace
		& g_t(x_t,u_t) + s_t = 0\,, \quad s_t \ge 0
	\end{aligned}
\end{equation}
where $V_{t+1}$ is a value function for the next state and $s_t = [s^1,\dots,s^k]^\top_t\in\mathbb{R}^k$ are slack variables. At the final stage, the value function is defined as $V_T(x_T) = l_f(x_T)$.

For notational convenience, we drop the time index $t$ in the remainder of this section, with the understanding that all functions and variables remain time-dependent. The relaxed Lagrangian with the log-barrier terms of $s$ is defined by the following $Q$-function:
\begin{equation}
	\label{eq:lagrangian}
		Q = l(x,u) + V'\!(f(x,u)) + y^\top\! (g(x,u) + s) - \mu \sum_{i=1}^{k} \log s^i
\end{equation}
where $\mu > 0$ is the barrier parameter and $y$ is the Lagrangian multiplier. The relaxed value function $V(x)$ is defined by a saddle point of the $Q$-function:
\begin{equation}\label{eq:valuefunc}
	V(x) = \underset{u,s}{\min} \ \underset{y}{\max} \enspace Q(x,u,s,y)
\end{equation}
	
\subsubsection{Backward Pass}
\label{sec:trajgen:ipddp:back}
As in the standard DDP scheme, $Q$ is perturbed up to the quadratic terms at the current nominal points:
%
\begin{align}
	\delta Q 
	&=
	\begin{bmatrix}
		Q_x\\Q_u\\Q_s\\Q_y
	\end{bmatrix}^{\!\!\top}  \!\!\!
	\begin{bmatrix}
		\delta x \\ \delta u \\ \delta s \\ \delta y
	\end{bmatrix} \!
	+
	{1 \over 2} \!
	\begin{bmatrix}
		\delta x\\\delta u\\\delta s\\\delta y
	\end{bmatrix}^{\!\!\top}  \!\!\! 
	\left[
	\begin{array}{@{\,}c@{\,\,}c@{\,\,}c@{\,\,}c@{\,}}
		Q_{xx} & Q_{xu} & Q_{xs} & Q_{xy}\\
		Q_{ux} & Q_{uu} & Q_{us} & Q_{uy}\\
		Q_{sx} & Q_{su} & Q_{ss} & Q_{sy}\\
		Q_{yx} & Q_{yu} & Q_{ys} & Q_{yy}\\
	\end{array}
	\right] \!
	\begin{bmatrix}
		\delta x\\\delta u\\\delta s\\\delta y
	\end{bmatrix}\nonumber\\
	&= 
	Q_x^\top \delta x + {1\over 2}\delta x^\top  Q_{xx}\delta x
	\label{eq:qfunc} \\
	& \quad 
	+ 
	\delta x^\top \! 
	\begin{bmatrix}
		Q_{xu} &\!\!\!\! 0 &\!\!\!\!  Q_{xy}
	\end{bmatrix} \! 
	\begin{bmatrix}
		\delta u \\ \delta s \\ \delta y
	\end{bmatrix}
	+
	\begin{bmatrix}
		Q_u\\y-\mu S^{-1}e\\Q_y
	\end{bmatrix}^{\!\!\top}  \!\!\! 
	\begin{bmatrix}
		\delta u\\\delta s\\\delta y
	\end{bmatrix}\nonumber\\
	& \quad 
	+
	{1 \over 2}
	\begin{bmatrix}
		\delta u\\\delta s\\\delta y
	\end{bmatrix}^{\!\!\top}  \!\!\! 
	\left[
	\begin{array}{@{\,}c@{\,\,}c@{\,\,}c@{\,}}
		Q_{uu} & 0 & Q_{uy}\\
		0 & \mu S^{-2} & I\\
		Q_{yu} & I & 0\\
	\end{array}
	\right]
	\!\!
	\begin{bmatrix}
		\delta u\\\delta s\\\delta y
	\end{bmatrix}\nonumber
\end{align}
%
where $e \in \mathbb{R}^k$ is an all-ones vector and $S:=\text{diag}(s) \in \mathbb{R}^{k\times k}$ is a diagonal matrix associated with the vector $s \in \mathbb{R}^k$. 
By setting $\delta s^\top (\mu S^{-2})\delta s = \delta s^\top ( S^{-1}Y)\delta s$ where $Y:=\text{diag}(y)$, the step direction that satisfy the extremum condition corresponding to the first-order optimality is determined by the following primal-dual KKT system:
\begin{equation}
	\label{eq:kktsys}
	\begin{aligned}
		\left[
		\begin{array}{@{\,}c@{\,\,\,}c@{\,\,\,}c@{\,}}
			Q_{uu} & 0 & Q_{uy}\\
			0 & Y & S\\
			Q_{yu} & I & 0\\
		\end{array}
		\right] \!\!
		\begin{bmatrix}
			\delta u\\\delta s\\\delta y
		\end{bmatrix}
		=
		-\begin{bmatrix}
			Q_{ux} \\ 0 \\ Q_{yx}
		\end{bmatrix}
		\delta x
		-
		\begin{bmatrix}
			Q_u\\Sy-\mu e\\Q_y
		\end{bmatrix}
	\end{aligned}
\end{equation}

Solving the KKT system~\eqref{eq:kktsys} for $\delta u,\delta s,\delta y$, we obtain
\begin{equation}
	\begin{aligned}
		\delta u = K_u\delta x + d_u \,, \, 
		\delta s = K_s\delta x + d_s \,, \,
		\delta y = K_y\delta x + d_y
	\end{aligned}
\end{equation}
where the coefficient matrices and vectors are defined as 
\begingroup
\allowdisplaybreaks
\begin{align} \label{bp}
	&K_u = -\tilde{Q}_{uu}^{-1}\tilde{Q}_{ux}\,,
	&&\!\! d_u = -\tilde{Q}_{uu}^{-1}\tilde{Q}_u\,,\nonumber\\
	&K_s = -(Q_{yx} + Q_{yu}K_u)\,,
	&&\!\! d_s = -(r_p + Q_{yu}d_u)\,, \\
	&K_y = S^{-1}Y(Q_{yx} + Q_{yu}K_u)\,,
	&&\!\! d_y = S^{-1}(r + YQ_{yu}d_u)  \nonumber
\end{align}
\endgroup
with the intermediate parameters and vectors
\begingroup
\allowdisplaybreaks
\begin{align} \label{bp2}
	&\tilde{Q}_u = Q_u + Q_{uy}S^{-1}r \,,\,
	&&\!\! r = Yr_p - r_d\,,\nonumber\\
	&\tilde{Q}_{uu} = Q_{uu} + Q_{uy}S^{-1}YQ_{yu}\,,\,
	&&\!\! r_p = Q_y\,,\\
	&\tilde{Q}_{ux} = Q_{ux} + Q_{uy}S^{-1}YQ_{yx}\,,\,
	&&\!\! r_d = Sy - \mu e \,.    \nonumber
\end{align}
\endgroup
Here, $r_p$ and $r_d$ are known as the primal and dual residuals, respectively.
The KKT variables $\delta s$ and $\delta y$ can be rewritten as 
\begin{equation}
	\begin{aligned}
		\delta s &= S^{-1}YQ_{yx}\delta x + S^{-1}YQ_{yu}\delta u + S^{-1}r\,, \\
		\delta y &= -Q_{yx}\delta x - Q_{yu}\delta u - r_p \,.
	\end{aligned}
\end{equation}
Substituting $\delta s, \delta y$ above into the quadratic form $\delta Q$ in (\ref{eq:qfunc}) and setting  $\delta s^\top (\mu S^{-2})\delta s = \delta s^\top ( S^{-1}Y)\delta s$ result in another representation for the perturbed quadratic form:
\begin{equation}
	\begin{aligned}
		\delta Q =& {1\over 2} r_p^\top S^{-1}Yr_p - r_d^\top S^{-1}r_p + 
		\begin{bmatrix}
			\tilde{Q}_x\\\tilde{Q}_u
		\end{bmatrix}^\top 
		\begin{bmatrix}
			\delta x\\\delta u
		\end{bmatrix}\\
		&+
		{1 \over 2}
		\begin{bmatrix}
			\delta x\\\delta u
		\end{bmatrix}^\top 
		\begin{bmatrix}
			\tilde{Q}_{xx} & \tilde{Q}_{ux}^\top \\
			\tilde{Q}_{ux} & \tilde{Q}_{uu}
		\end{bmatrix}
		\begin{bmatrix}
			\delta x\\\delta u
		\end{bmatrix}
	\end{aligned}
\end{equation}
where $\tilde{Q}_x = Q_x + Q_{xy}S^{-1}r$ and $\tilde{Q}_{xx} = Q_{xx} + Q_{xy}S^{-1}YQ_{yx}$.

Finally, we obtain the perturbed value function as follows:
\begin{equation}
	\label{eq:value}
	\delta V = \underset{\delta u,\delta s}{\min}\underset{\delta y}{\max}\enspace \delta Q = \Delta V + V_x^\top \delta x + {1\over 2}\delta x^\top  V_{xx} \delta x
\end{equation}
where the coefficients are given as
	\begin{align}
		\Delta V &= \tilde{Q}^\top _u d_u + {1\over 2} d_u^\top \tilde{Q}_{uu}d_u + {1\over 2} r_p^\top S^{-1}Yr_p - r_d^\top S^{-1}r_p \,, \nonumber\\
		V_x &= \tilde{Q}_x + K_u^\top \tilde{Q}_u + \tilde{Q}_{ux}^\top d_u + K_u^\top \tilde{Q}_{uu} d_u \,, \\
		V_{xx} &= \tilde{Q}_{xx} + K_u^\top \tilde{Q}_{ux} + \tilde{Q}_{ux}^\top K_u + K_u^\top \tilde{Q}_{uu} K_u \,. \nonumber
	\end{align}
This perturbed value function $\delta V$ is recursively used for $\delta V^\prime$ at the next backward step.

\subsubsection{Forward Pass}
\label{sec:trajgen:ipddp:forward}
After calculating the perturbations in the backward pass, the nominal points are updated as follows:
$u \leftarrow u + \alpha\delta u \enspace\!\!,\enspace\!
s \leftarrow s + \alpha\delta s \enspace\!\!,\enspace\!
y \leftarrow y + \alpha\delta y$
where $\alpha\in(0,1]$  represents the step size. In IPDDP, the value of $\alpha$ is determined by the filter line-search method \cite{wachter2006implementation}. This method starts with a step size of 1 and reduces $\alpha$ incrementally. pdates are accepted as soon as they decrease either the cost or the violations of constraints. If no suitable $\alpha$ is found, the forward pass is terminated and deemed unsuccessful.
	
\subsubsection{Convergence}
\label{sec:trajgen:ipddp:converge}
The barrier parameter $\mu$ is monotonically decreased whenever the local convergence to the central path has been achieved. The criterion for the local convergence is $\max(\|Q_u\|_{\infty},\|r_p\|_{\infty},\|r_d\|_{\infty}) < \kappa\mu$ for some $\kappa > 1$. The global convergence agrees with the sufficiently small $\mu$.
	
\subsubsection{Regularization}
\label{sec:trajgen:ipddp:regular}
To guarantee that $\tilde{Q}_{uu}^{-1}$ is invertible in (\ref{bp}), the regularization parameter $\rho \ge 0$ is added: $Q_{uu} \leftarrow Q_{uu} + \rho I_{m}$. The parameter $\rho$ increases when it is not invertible or the failure has occurred in the forward pass. If $\rho$ reaches some upper bound $\rho_{\max}$, IPDDP is terminated for failure.

\section{Collision-free Smooth Trajectory Generation}\label{sec:method}
This section considers the following OCP:
\begin{equation}\label{eq:ocp_mppi_ddp}
	\begin{aligned}
		\underset{u_{0:T-1}}{\text{minimize}} \quad &l_f(x_T) + \sum_{t=0}^{T-1} l(x_t,u_t)\\
		\text{subject to} \quad 
		& x_{t+1} = f_t(x_t,u_t),\,x_0 = x_{\rm init}\\
		& g_{t}^{x}(x_t) \leq 0, \ g_{t}^{u}(u_t) \leq 0\,, \quad p_t \notin \mathcal{O}
	\end{aligned}
\end{equation}
where the variables $x_t \in \mathbb{R}^n$ and $u_t \in \mathbb{R}^m$ are the system state and the control input vector at time-step $t$, respectively, and $x_{\rm init} $ is the initial condition for the control system. $p_t \in x_t$ is the position of a robot, and $\mathcal{O}$ is the set of positions at which obstacles occupy. The functions $l_f : \mathbb{R}^{n} \rightarrow \mathbb{R}$, $l_t: \mathbb{R}^{n} \times \mathbb{R}^{m} \rightarrow \mathbb{R}$, and $f_t: \mathbb{R}^{n} \times \mathbb{R}^{m} \rightarrow \mathbb{R}^{n}$ are defined as~\eqref{eq:ocp}. Notice that, unlike in~\eqref{eq:ocp}, the joint state-control constraints \( g_t(x_t, u_t) \leq 0 \) are decoupled into the state constraint \( g_t^x(x_t) \leq 0 \) and the input (control) constraint \( g_t^u(u_t) \leq 0 \).

The proposed algorithm for solving (\ref{eq:ocp_mppi_ddp}) has three steps: searching for a feasible coarse trajectory using MPPI, constructing path corridors, and smoothing the coarse trajectory by IPDDP.

\subsection{Model Predictive Path Integral}
\label{sec:method:mppi}
We first generate a coarse trajectory using MPPI. The cost function $J$ is defined as
\begin{equation}\label{eq:mppi}
	J(U) = l_f(x_T) + \sum_{t=0}^{T-1} l_t(x_t,u_t) + \mathcal{I}^\text{\scriptsize MPPI}(x_t)
\end{equation}
where the indicator function is defined as
\begin{equation}\label{eq:collision}
		\mathcal{I}^\text{\scriptsize MPPI}(x_t) =
		\begin{cases}
			\infty, & \text{if\enspace} p_t\in\mathcal{O} 
			\text{\enspace or\enspace} g_t^{x}(x_t) > 0\\
			0, & \text{otherwise}
		\end{cases}
	\end{equation}
ensuring obstacle avoidance and the sequence of the states $x_{0:T}$ are determined by the initial state $x_0 = x_{\rm init}$, the dynamics $x_{t+1} = f_{t}(x_t,u_t)$, and the controls $U$. 
	
To satisfy the control constraints in (\ref{eq:ocp_mppi_ddp}), each $i$th sample of control sequence vector $U_i$ is projected onto the constraint set, i.e. $U_i \leftarrow \Pi_{u}(U_i)$ where $\Pi_{u}$ is a projection operator onto the feasible set of controls $\mathcal{U} = \{u_{0:T-1}\in \mathbb{R}^{mT} \,|\, g_{t}^{u}(u_t)\le 0 \ \text{for all  $t$}\}$. We assume that the set \( \mathcal{U} \) is compact and convex, ensuring that the projection is well-defined. This assumption allows us to leverage analytical solutions for projection, particularly in cases involving simple constraints like box constraints or second-order conic constraints. 

With the method described in Section \ref{sec:preliminaries:viso:traj}, locally optimal controls and corresponding states are obtained. Let $\bar{p}_{0:T}$ be the resulting position of a robot from MPPI.

\vspace{-4mm}
\subsection{Path Corridors}
\label{sec:method:corridor}
\begin{figure}[b]\vspace{-6mm}
	\centering
	\subfloat[A maximally inflated path corridor]{\includegraphics[width=0.485\linewidth,height=45.5mm]{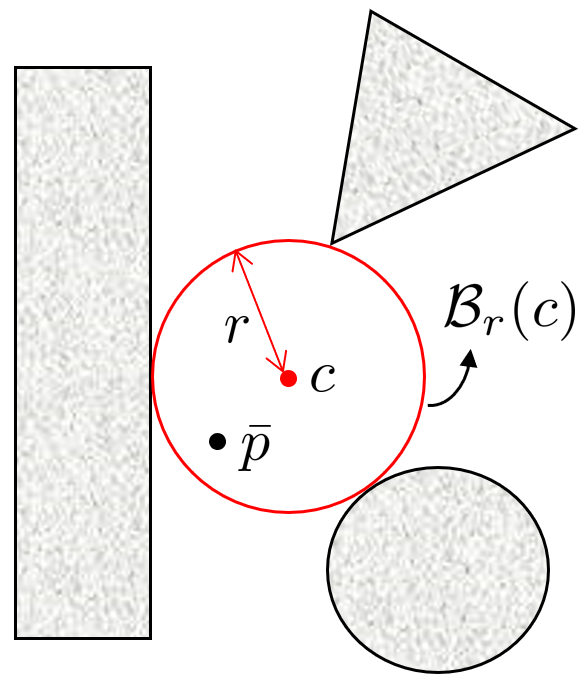}}\hspace{2mm}
	\subfloat[Inflation process]{\includegraphics[width=0.35\linewidth,height=50mm]{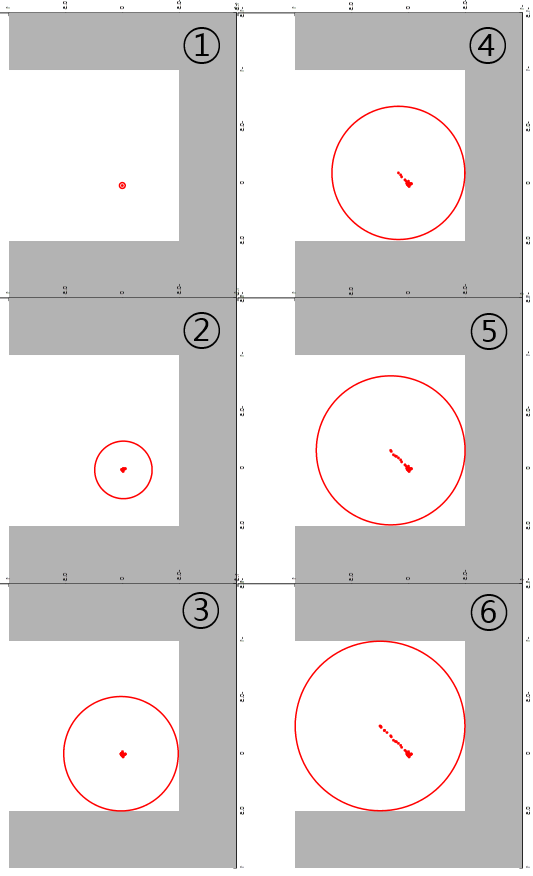}}\vspace{-1mm}
	\caption{Schematics of for collision-free path corridors.}
	\label{fig:corridor}
\end{figure}

To construct corridors around the path $\bar{p}_{0:T}$, the following optimization problem is considered:
\begin{equation}\label{eq:corridor}
	\begin{aligned}
		\underset{c,r}{\text{minimize}} \quad &J(c,r) = \lambda_c\|c - \bar{p}\|_2 - \lambda_rr + \mathcal{I}^\text{\scriptsize PC}(c,r)\\
		\text{subject to} \quad 
		& 0 \le r \le r_{\rm max}
	\end{aligned}
\end{equation}
where the indicator function for a radial collision-free corridor is defined as
\begin{equation}\label{eq:collision2}
	\begin{aligned}
		\mathcal{I}^\text{\scriptsize PC}(c,r) &=
		\begin{cases}
			\infty, & \text{if\enspace} \mathcal{B}_r(c) \bigcap \mathcal{O} \neq \emptyset\\
			0, & \text{otherwise} 
		\end{cases} \,, \\[2pt]
		\mathcal{B}_r(c) &= \{ p \ | \ \|p - c\|_2 \le r \}\,.
	\end{aligned}
\end{equation}
The parameters $\lambda_c,\lambda_r > 0$ are the weights, and $r_{\max}$ is the maximum value of $r$. Although the shape of the corridors can be arbitrary, here we choose a  Euclidean ball $\mathcal{B}_r(c)$ which is represented by two variables: center $c$ and radius $r$. The optimization problem (\ref{eq:corridor}) is designed to enlarge the ball and have the center $c$ close to $\bar{p}$ while containing $\bar{p}$ inside the ball without intersection with obstacles (see Fig.~\ref{fig:corridor}). If there are no obstacles around $\bar{p}$, then the solution is $c = \bar{p}$ and $r = r_{\max}$.
	
We use the method described in Section~\ref{sec:preliminaries:viso:vector} with $\theta = [c^\top,r]^\top$ to solve the optimization problem~(\ref{eq:corridor}) at each stage of path planning to compute a sequence of collision-free corridors that are represented by $C = [c_0^\top,\dots,c_{T-1}^\top]^\top$ and $R = [r_0,\dots,r_{T-1}]^\top$. As in MPPI, the constraints on $r$ in~\eqref{eq:corridor} can be met by projection that is defined as $r \leftarrow \Pi_{z}(r) = \min\{r_{\max}, \max\{0,r\}\}$.

\vspace{-4mm}
\subsection{Trajectory Smoothing}
\label{sec:method:smooth}
In our final step of trajectory optimization for path planning, we consider the following OCP for smoothing the coarse trajectory generated by MPPI:
\begin{equation}\label{eq:ddp}
	\begin{aligned}
		\underset{u_{0:T-1}}{\text{minimize}} \quad &l_f(x_T) + \sum_{t=0}^{T-1}\! \left(  l_t(x_t,u_t) + \|p_t - c_t\|^2_Q \right) \\
		\text{subject to} \quad 
		& x_{t+1} = f_t(x_t,u_t), \,x_0 = x_{\rm init} \\
		& g_{t}^{x}(x_t) \leq 0, \ g_{t}^{u}(u_t) \leq 0, \,  \|p_t - c_t\|_2 \leq r_t \\
	\end{aligned}
\end{equation}
where $p_t\in x_t$ is, again, the position of a robot, $(c_t, r_t)$ are the center and radius of the path corridor computed in~\eqref{eq:corridor}, and $Q$ is a weight matrix penalizing deviations from the center of the corridor. We include the constraint in the last row of (\ref{eq:ddp}) to to keep a robot staying inside the collision-free corridors.

We use IPDDP introduced in Section~\ref{sec:preliminaries:alddp} to solve (\ref{eq:ddp}) and obtain a smooth trajectory.
At the time, the coarse trajectory from MPPI can be used for an initial guess, i.e., a warm start for local optimization, which can much accelerate the convergence of IPDDP. 

\begin{algorithm}[bh]\small
	\caption{${\tt MPPI\mbox{-}IPDDP}$}\label{alg:MPPI-DDP}
	\begin{algorithmic}[1]\small
		\State \textbf{Input:} initial state $x_0$, collision checker
		\State \textbf{Output:} locally optimal controls $U^*$
		\State \textbf{Initialize:} controls $U$
		\While{not terminated}
		\State $U \leftarrow {\tt MPPI}(x_0,U)$
		\For{$t = 0,1,\dots,T-1$}
		\State $x_{t+1} \leftarrow f_{t} (x_t,u_t)$
		\EndFor
		\State $(C,R) \leftarrow {\tt Corridor}(X)$ 
		\Comment{$X = [ x^\top_0,\dots,x^\top_T]^\top$}
		\State $U \leftarrow {\tt IPDDP}(X,U,C,R)$ \Comment{(\ref{eq:ddp})}
		\EndWhile
	\end{algorithmic}
\end{algorithm}
\vspace{-4mm}
\begin{algorithm}[bh]\small
	\caption{${\tt MPPI}$}\label{alg:MPPI}
	\begin{algorithmic}[1]
		\For{$i=1,\dots,N_u$}
		\State $\hat{U}_i \leftarrow \Pi_u(U + \varepsilon_i) \ , \quad \varepsilon_i\sim\mathcal{N}(0,\Sigma_u)$
		\State $J_i \leftarrow {\tt cost}(x_0,\hat{U}_i)$
		\Comment{(\ref{eq:mppi})}
		\EndFor
		
		\State $\bar{J} \leftarrow \min_{i} J_i$
			
		\For{$i=1,\dots,N_u$}
		\State $J_i \leftarrow J_i - \bar{J}$
		\State $w_i \leftarrow \exp(-\gamma_u J_i)$ 
		\EndFor
			
		\State $\bar{w} \leftarrow \sum_{i=1}^{N_u} w_i$
		\State $U \leftarrow \sum_{i=1}^{N_u}(w_i/\bar{w}) \hat{U}_i$
		\State $U \leftarrow \Pi_u(U)$
	\end{algorithmic}\
\end{algorithm}\vspace{-2mm}

\subsection{Algorithms} 
\label{sec:method:algorithm}
Algorithm \ref{alg:MPPI-DDP} outlines the proposed trajectory optimization method, named {MPPI-IPDDP}, which is designed to generate collision-free, smooth trajectories. The algorithm includes three subroutines. First, {\tt MPPI} employs a derivative-free variational inference approach to search for a dynamically feasible but coarse trajectory. Second, {\tt Corridor} also utilizes derivative-free variational inference to construct collision-free circular corridors around the coarse trajectory. Lastly, {\tt IPDDP} uses a recursive method to smooth the coarse trajectory within these corridors. As demonstrated in the supplementary video, the proposed {\tt MPPI-IPDDP} method has been verified to be capable of online replanning for low-speed robots.

%
%
%
%
%
%
\begin{algorithm}[!t]\small
    \caption{${\tt Corridor}$}\label{alg:Corridor}
    \begin{algorithmic}[1]\small \vspace{-2mm}
	\State $\bar{p}_{0:T} \leftarrow \text{extract positions from $X$}$
	\State $c_t \leftarrow \bar{p}_t \enspace \text{for every $t=0,\dots,T-1$}$ 
	\State $r_t \leftarrow 0 \enspace \text{for every $t=0,\dots,T-1$}$
	\State $z_t \leftarrow [c^\top_t,r_t]^\top \enspace \text{for every $t=0,\dots,T-1$}$ 
        \For{$t=0,\dots,T-1$}
            \While{not inflated enough}
            \For{$i=1,\dots,N_z$}
            \State $\hat{z}_i \leftarrow \Pi_z(z_t + \varepsilon_i) \ , \quad \varepsilon_i\sim\mathcal{N}(0,\Sigma_z)$
            \State $J_i \leftarrow {\tt cost}(\hat{z}_i)$
            \Comment{(\ref{eq:corridor})}
            \EndFor
            
            \State $\bar{J} \leftarrow \min_{i} J_i$
                
            \For{$i=1,\dots,N_z$}
            \State $J_i \leftarrow J_i - \bar{J}$
            \State $w_i \leftarrow \exp(-\gamma_z J_i)$ 
            \EndFor
            
            \State $\bar{w} \leftarrow \sum_{i=1}^{N_z} w_i$
            \State $z_t \leftarrow \sum_{i=1}^{N_z}(w_i/\bar{w}) \hat{z}_i$
            \State $z_t \leftarrow \Pi_z(z_t)$
            \EndWhile
        \EndFor 
        \State $(C,R) \leftarrow [z_0^\top,z_1^\top,\dots,z_{T-1}^\top]^\top$ \quad (unpacking)
    \end{algorithmic}
\end{algorithm}
\vspace{-4mm}
\begin{algorithm}[!t]\small
	\caption{${\tt IPDDP}$}\label{alg:IPDDP}
	\begin{algorithmic}[1]\small
		\While{not converged globally and not max iteration}
		\State evaluate all derivatives needed;
		\State try the backwardpass;
		\Comment{Section~\ref{sec:trajgen:ipddp:back}}
		\State try the forwardpass;
		\Comment{Section~\ref{sec:trajgen:ipddp:forward}}
		
		\If{any failures occured} \Comment{Section~\ref{sec:trajgen:ipddp:regular}}
		\State increase the regularization parameter $\rho$;
		\If{$\rho > \rho_{\max}$}
		\State break;
		\Comment{Solver failed}
		\EndIf
		\State continue;
		\Else
		\State decrease the regularization parameter $\rho$;
		\State update the nominal trajectory;
		\EndIf
		
		\If{locally converged} \Comment{Section~\ref{sec:trajgen:ipddp:converge}}
		\State decrease the barrier parameter $\mu$;
		\State reinitialize the filter;
		\EndIf
		
		\EndWhile
	\end{algorithmic}
\end{algorithm}

\section{Case Studies}\label{sec:case}

\subsection{Wheeled Mobile Robot}
\label{sec:case:mobile}
\begin{figure*}[h!]
	\centering
	\includegraphics[width=.83\textwidth, height=55mm]{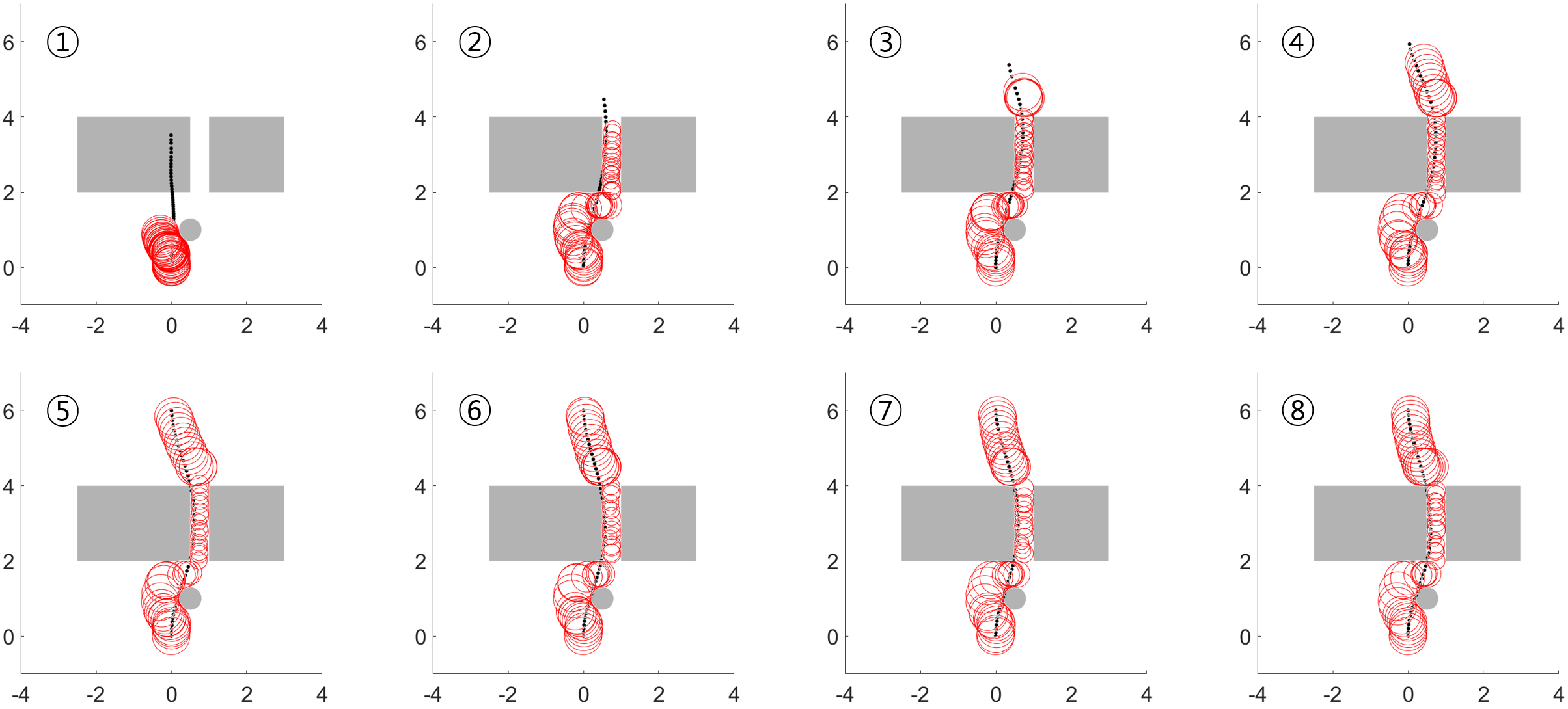}\vspace{-3mm}
	\caption{The iterations of the {\tt MPPI-IPDDP} algorithm for generating a collision-free path from $(0,0)$ to $(0,6)$. In the figure, black dots represent the positions of the robot, red circles denote the path corridors, and gray areas indicate obstacles. During the early iterations, the constraints are violated (as the black dots are outside the corridors) because the {\tt IPDDP} struggled with the infeasible starting point and was terminated by the user-defined maximum iteration limit, as illustrated in \textcircled{1}$\sim$\textcircled{3}. However, as the {\tt MPPI-IPDDP} algorithm continues to iterate, it eventually finds the optimal collision-free trajectory, as shown in \textcircled{8}.}
	\label{fig:optseq}\vspace{-6mm}
\end{figure*}
\begin{figure}[t]
	\centering
	\subfloat[Coarse controls by {\tt MPPI}]{\includegraphics[width=0.425\linewidth, height=28mm]{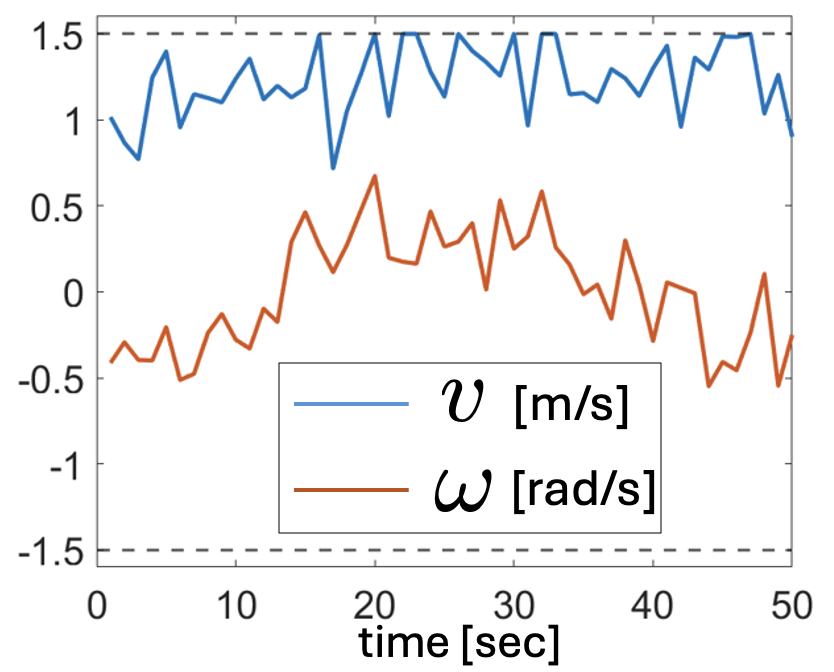}}\hspace{1mm}
	\subfloat[Smooth controls by {\tt IPDDP}]{\includegraphics[width=0.425\linewidth, height=28mm]{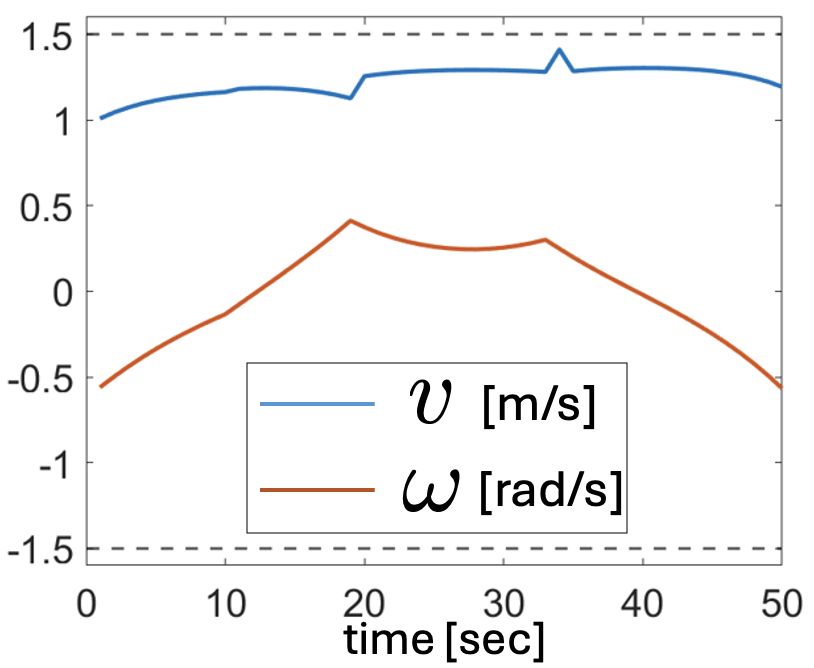}}\vspace{-1mm}
	\caption{A comparison of control inputs obtained from {\tt MPPI} and {\tt IPDDP}.}
	\label{fig:mobilecontrol}
\vspace{-1mm}
	\centering
	\subfloat[The terminal state cost of the trajectory reduces over iterations.]{\includegraphics[width=0.425\linewidth, height=28mm]{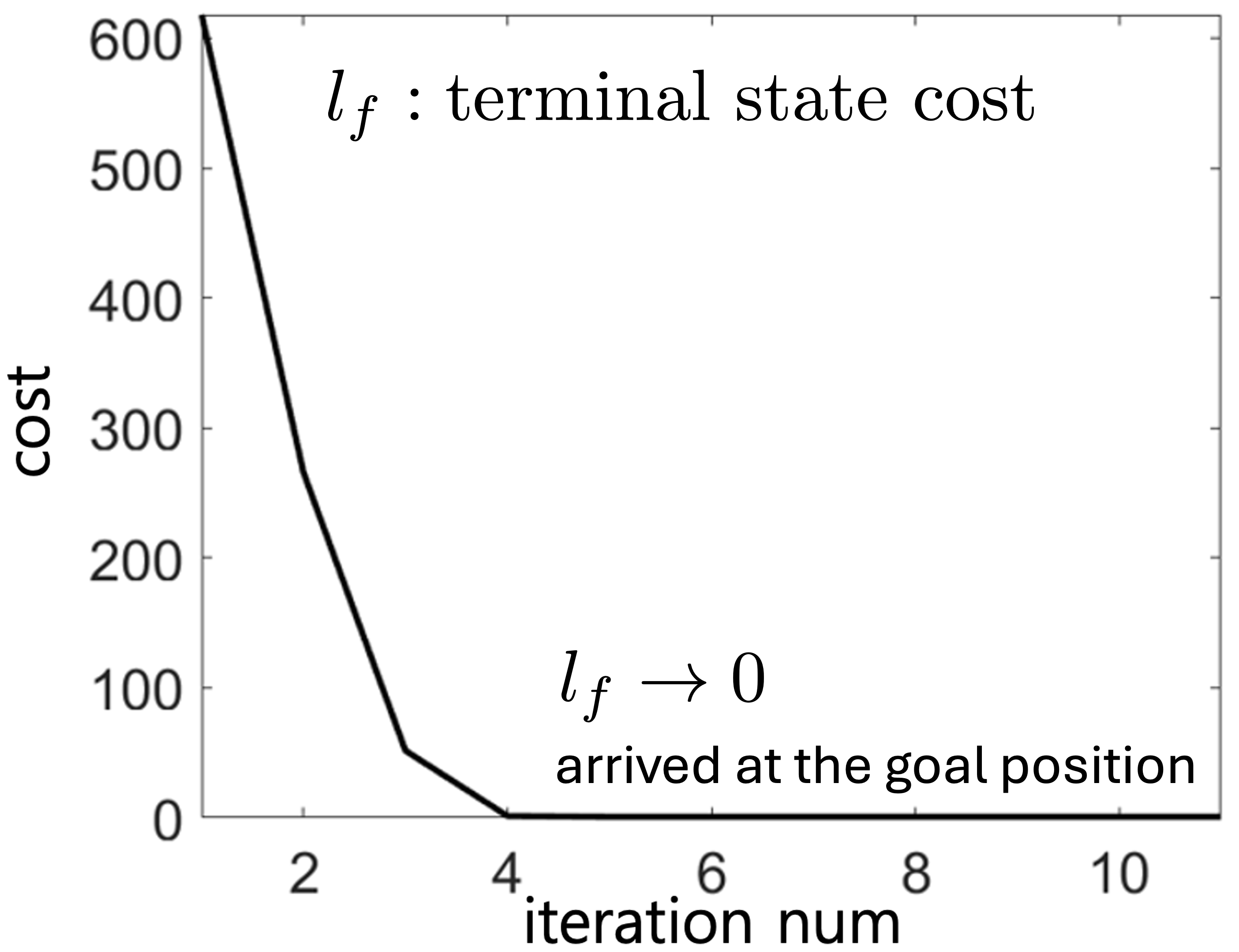}}\hspace{1mm}
	\subfloat[The maximum value of the primal residuals approaches 0, meaning that the constraints are satisfied.]{\includegraphics[width=0.455\linewidth, height=28mm]{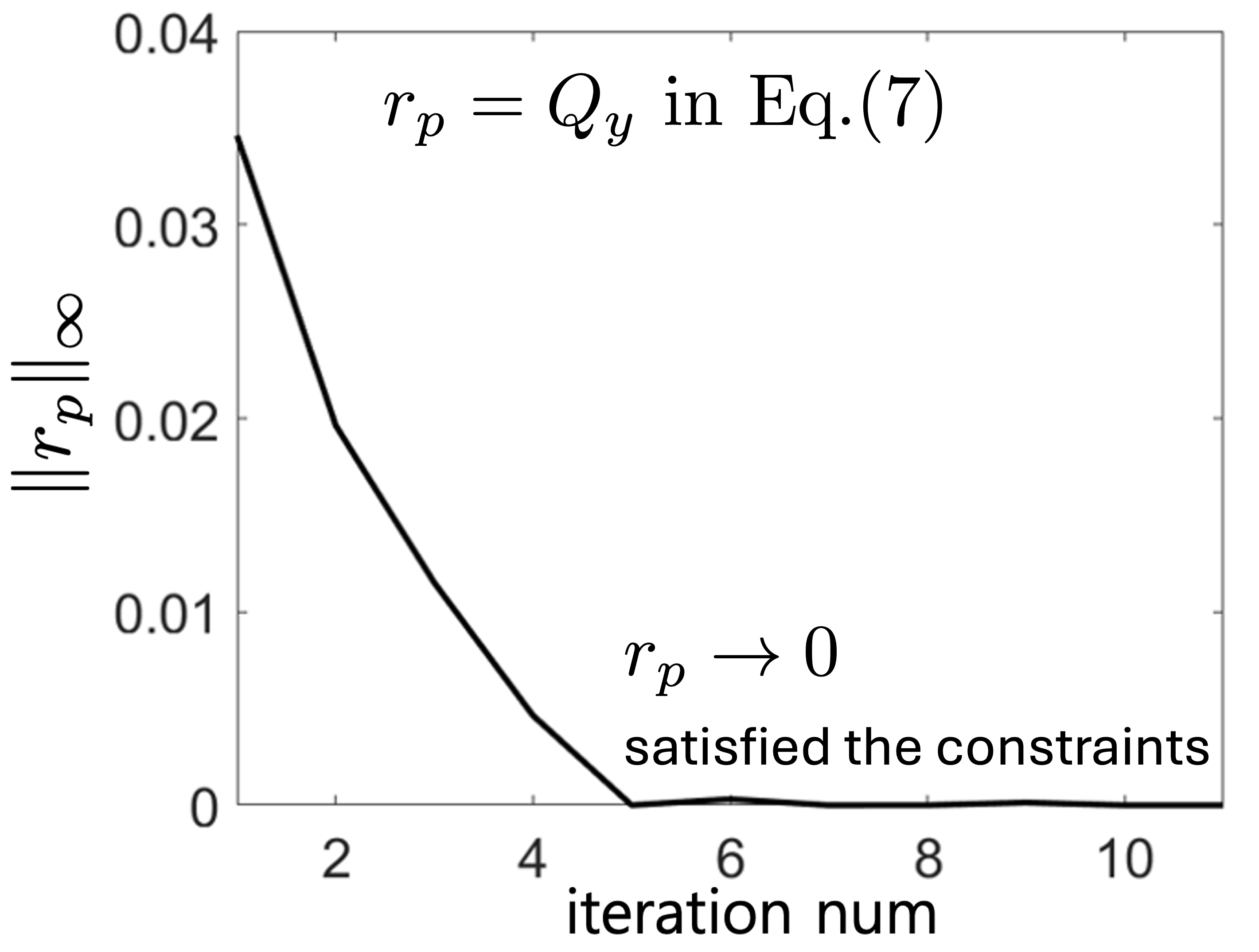}}\vspace{-1mm}
	\caption{Cost reduction and convergence rate over {\tt MPPI-IPDDP} iterations.}
	\label{fig:cost}\vspace{-2mm}
\end{figure}
	
For an example of path planning in 2D space, we consider a scenario that a differential wheeled robot arrives at a given target pose without collision. Consider the robot kinematics\\[-1mm]
\begin{equation}
	\begin{split}
		x_{t+1} &= x_t + v_t\cos(\theta_t)\Delta t \,,\
		y_{t+1} = y_t + v_t\sin(\theta_t)\Delta t \,, \\
		\theta_{t+1} &= \theta_t + w_t\Delta t\\
	\end{split}
\end{equation}
where $(x_t,y_t)\in\mathbb{R}$ are the positions of the x-axis and y-axis respectively,  $\theta_t\in\mathbb{R}$ is the angle of the orientation, $v_t,w_t\in\mathbb{R}$ are velocity and angular velocity respectively, and $\Delta t$ is the time interval. The vectors $[x_t,y_t,\theta_t]^\top$ and $[v_t,w_t]^\top$ are states and controls respectively. We set the initial states as $[0,0,\pi/2]^\top$ and sampling-time interval $\Delta t = 0.1$. 
	
The constraints of the corresponding OCP for trajectory generation are defined as
\begin{equation}
	\begin{aligned}
		0 \le v_t \le 1.5\ , \ 
		|w_t| \le 1.5\ , \
		(x_t,y_t)\notin\mathcal{O}
	\end{aligned}
\end{equation}
where $\mathcal{O}$ is the set of obstacles shown in Fig.~\ref{fig:optseq} in gray. 
The cost functions of the corresponding OCP for trajectory generation are defined as
\begin{equation*}
l_f = 300(x_T^2 + (y_T\!- 6)^2 + (\theta_T - \pi/2)^2) , \,  l_{t} = 0.01(v_t^2 + w_t^2)
\end{equation*}
where $[0,6,\pi/2]^\top$ is the target pose. The parameters for the MPPI-IPDDP method are given in Tab.~\ref{tab:parametermobile}.

Fig.~\ref{fig:optseq} shows the processing results of generating a smooth trajectory. Fig.~\ref{fig:mobilecontrol} gives a comparison between the zigzaging controls obtained from MPPI and the smoother ones by IPDDP. Fig.~\ref{fig:cost} shows that the cost and constraint violations reduce over MPPI-IPDDP iterations.

\begin{table}[b]\vspace{-2mm}
	\caption{Parameters for trajectory optimization of a wheeled mobile robot in Section~\ref{sec:case:mobile}.}\vspace{-4mm}
	\begin{center}
		\renewcommand{\arraystretch}{1.05}
		\begin{tabular}{|c|c||c|c|}
			\hline
			\textbf{Parameter} & \textbf{Value} & \textbf{Parameter} & \textbf{Value}\\
			\hline
			$\lambda_c$ & 20 & 
			$\lambda_r$ & 35 \\
			\hline
			$r_{\text{max}}$ & 0.5 & 
			$T$ & 50 \\
			\hline
			$\gamma_z$ & 1000 & 
			$\gamma_u$ & 100 \\
			\hline
			$N_u$ & 5000 &
			$\Sigma_u$ & $\text{diag}([0.25, 0.25])$ \\
			\hline
			$N_z$ & 3000 &
			$\Sigma_z$ & $\text{diag}([0.3, 0.3, 0.08])$ \\
			\hline
			$Q$ & $0.001I_3$ & 
			&  \\
			\hline
		\end{tabular}
	\end{center}
	\vspace{-5mm}
	\label{tab:parametermobile}
\end{table}

\begin{table}[b]
	\caption{Parameters for trajectory optimization of a quadrator in Section~\ref{sec:case:quadrotor}.}\vspace{-4mm}
		\begin{center}
			\renewcommand{\arraystretch}{1.05}
			\begin{tabular}{|c|c||c|c|}
				\hline
				\textbf{Parameter} & \textbf{Value} & \textbf{Parameter} & \textbf{Value}\\
				\hline
				$\lambda_c$ & 20 & 
				$\lambda_r$ & 35 \\
				\hline
				$r_{\text{max}}$ & 0.5 & 
				$T$ & 30 \\
				\hline
				$\gamma_z$ & 1000 & 
				$\gamma_u$ & 100 \\
				\hline
				$N_u$ & 8000 &
				$\Sigma_u$ & $\text{diag}([1.5, 1.5, 1.5])$ \\
				\hline
				$N_z$ & 5000 &
				$\Sigma_z$ & $\text{diag}([0.3, 0.3, 0.3, 0.08])$ \\
				\hline
				$Q$ & $0.001I_3$ & 
				&  \\
				\hline
			\end{tabular}
		\end{center}
	\label{tab:parameterdrone}
\end{table}

\subsection{Quadrotor without Attitude}
\label{sec:case:quadrotor}

\begin{figure}[ht]
	\centering
	\includegraphics[width=.975\linewidth, height=58mm]{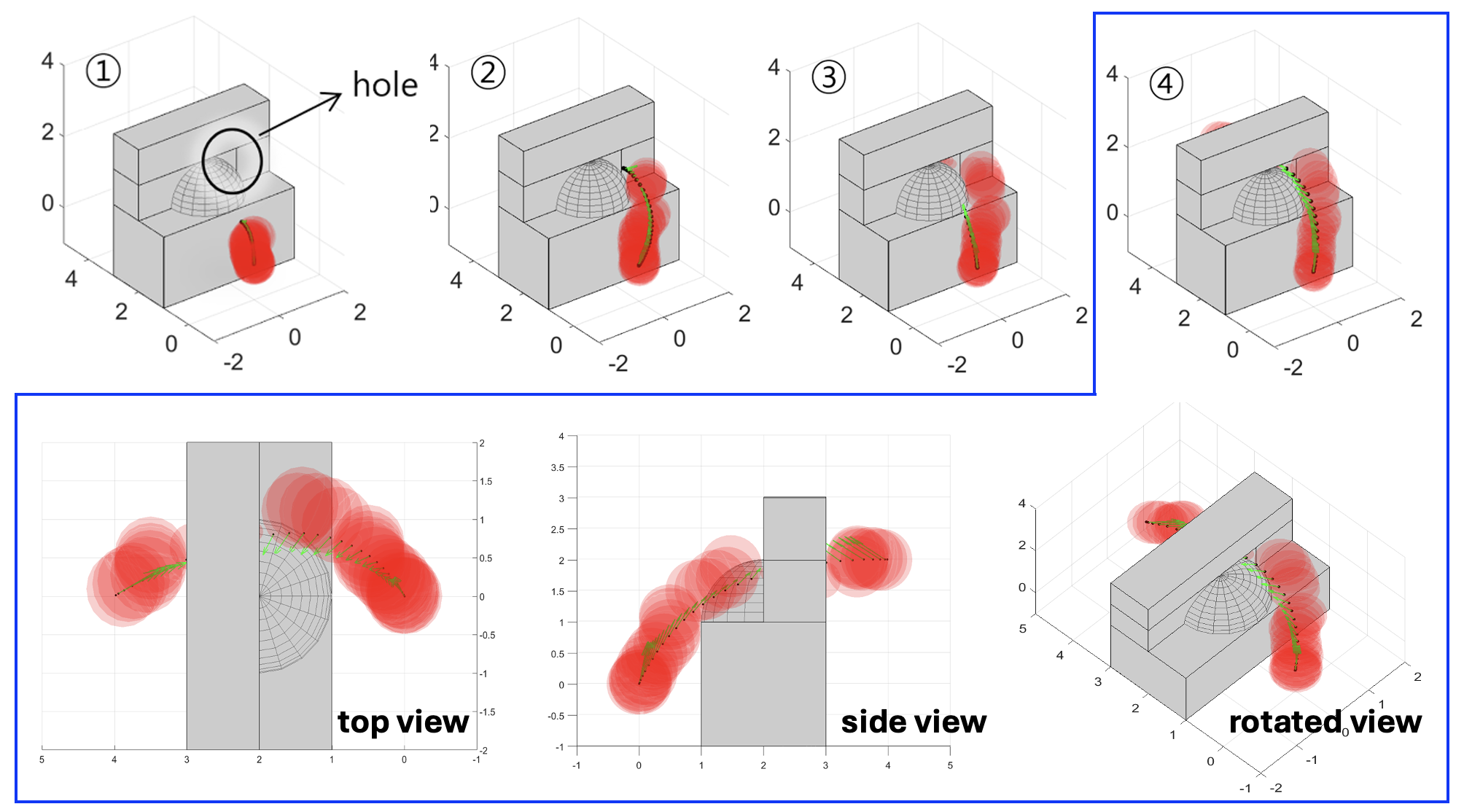}\vspace{-4mm}
	\caption{The iterations for generating an optimal collision-free path from $(0,0,0)$ to $(0,4,2)$ by {\tt MPPI-IPDDP}. The black dots are position of the quadrotor, the red spheres are the path corridors, and the gray represents obstacles. The optimal trajectory passes the small hole and reaches the destination. Details can be found in accompanying video.}
	\label{fig:optseqdrone}\vspace{-2mm}
\end{figure}


\begin{figure}[t!]\vspace{-4mm}
	\centering
	\subfloat[Coarse controls by {\tt MPPI}]{\includegraphics[width=0.425\linewidth, height=28mm]{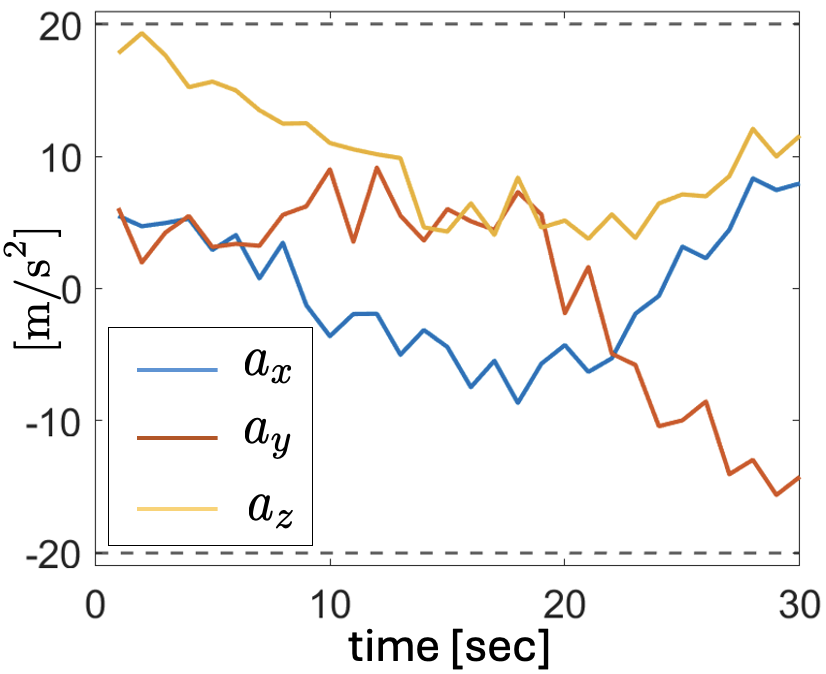}}\hspace{1mm}
	\subfloat[Smooth controls by {\tt IPDDP}]{\includegraphics[width=0.425\linewidth, height=28mm]{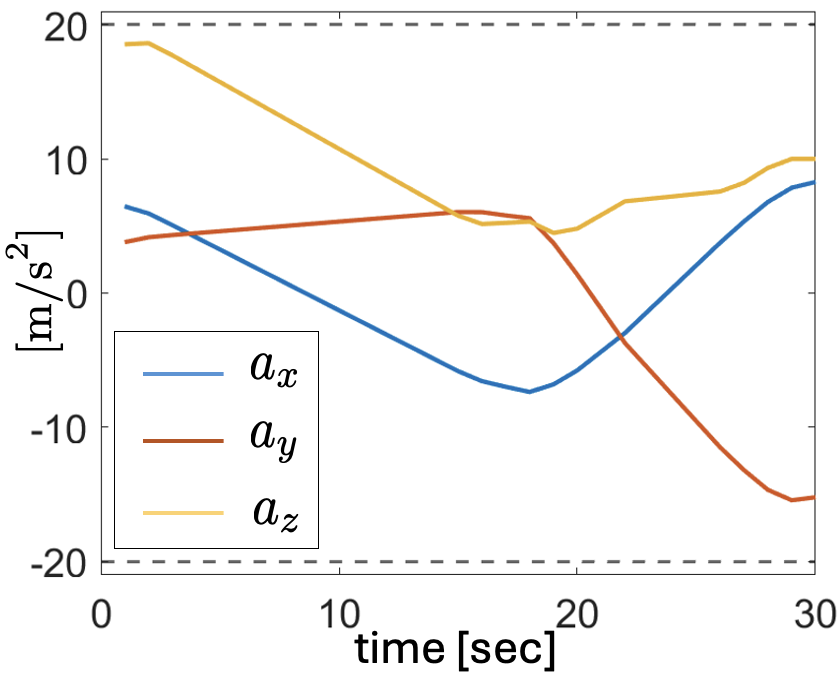}}\vspace{-1mm}
	\caption{A comparison of control inputs obtained from {\tt MPPI} and {\tt IPDDP}.}
	\label{fig:quadcontrol}
\vspace{-1mm}
	\centering
	\subfloat[{The terminal state cost of the trajectory reduces over iterations.}]{\includegraphics[width=0.45\linewidth, height=30mm]{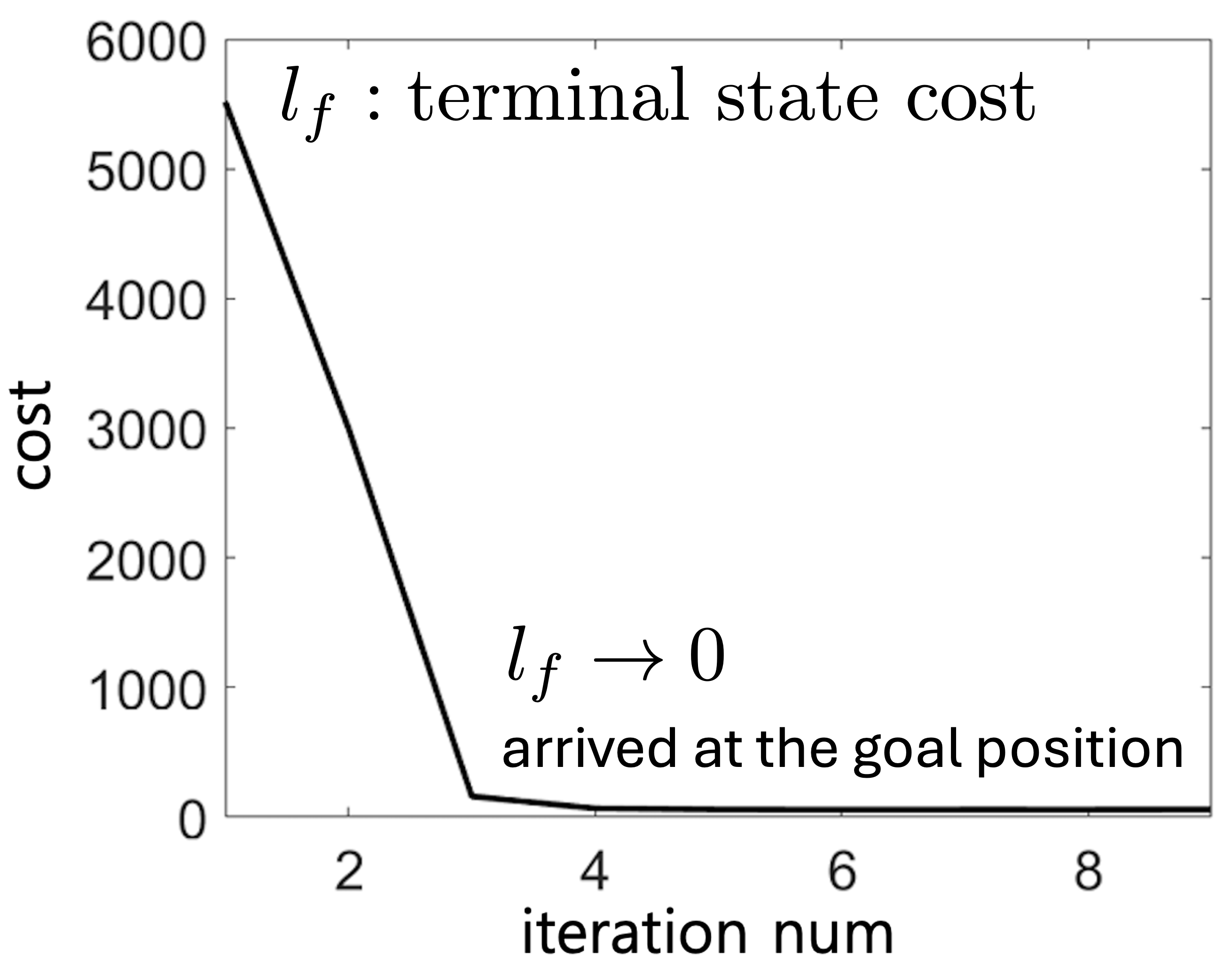}}\hspace{1mm}
	\subfloat[{The maximum value of the primal residuals approaches 0, meaning that the constraints are satisfied.}]{\includegraphics[width=0.455\linewidth, height=30mm]{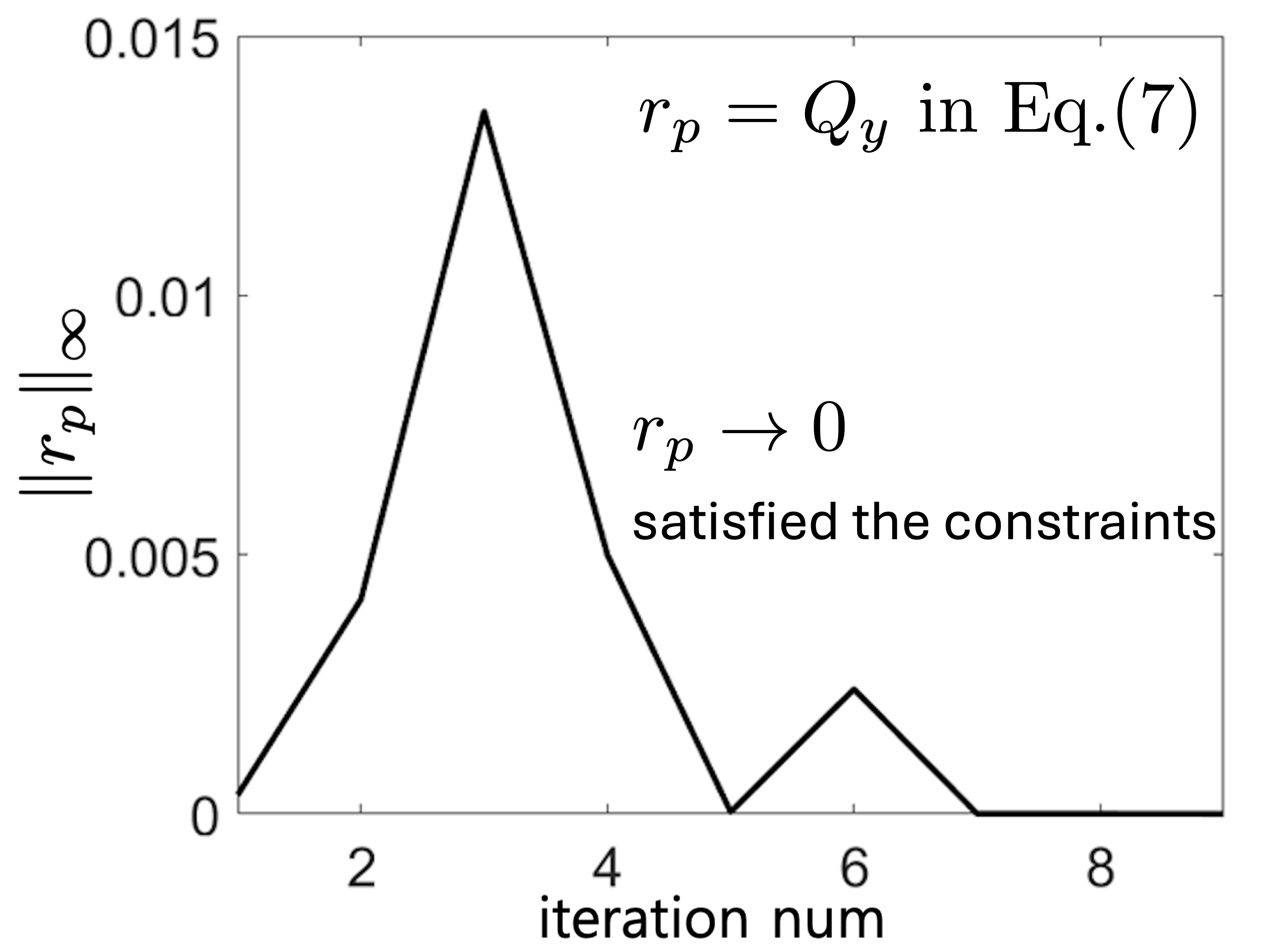}}\vspace{-1mm}
	\caption{Cost reduction and convergence rate over {\tt MPPI-IPDDP} iterations.}
	\label{fig:quadcost}\vspace{-2mm}
\end{figure}

For an example of path planning in 3D space, we consider a scenario that a quadrotor arrives at a given target position without collision.
We assume that the quadrotor can be modeled as a point mass. The kinematics is given by
	\begin{equation}
		\begin{aligned}
			x_{t+1} &= x_t + v_t\Delta t \,, \ 
			v_{t+1} = v_t + (a_t - ge_3)\Delta t\\
		\end{aligned}
	\end{equation}
where $x_t,v_t\in\mathbb{R}^3$ are position and velocity respectively, $a_t = [a_{x,t}, a_{y,t}, a_{z,t}]^\top \!\in\mathbb{R}^3$ is acceleration, $g=9.81$ is the gravitational acceleration, and $e_3 = [0,0,1]^\top$ is the vector of $z$-axis. $[x_t^\top,v_t^\top]^\top$ and $a_t$ are the state and control respectively. We set the initial state as $[0,0,0,0,0,0]^\top$ and $\Delta t = 0.05$.
	
The constraints of the corresponding OCP for trajectory generation are defined as $x_t\notin\mathcal{O}$ and $a_{t} \in \mathcal{K}$ with
\begin{equation}\label{eq:quadcons}
\!\!\mathcal{K} \!=\! {\{ \| u \|_2 \!\le\! a_{\max} \}}  \cap {\{ \|u\|_2\cos(\theta_{\max}) \!\le\! e_3^\top u \}} \!=\! \mathcal{K}_{1} \!\cap \mathcal{K}_{2}
\end{equation}
where $a_{\max} = 20$ 
and $\theta_{\max}=60^\circ$ are the maximum value of acceleration and thrust angle, respectively.
This ensures the acceleration vector of the quadrotor remain within a defined conic region $\mathcal{K}$, and $\mathcal{O}$ is the set of obstacles shown in Fig.~\ref{fig:optseqdrone} in gray. When the projection is performed to satisfy the conic constraint $\mathcal{K}_2$ in~\eqref{eq:quadcons}, we consider the following projection operator for the second-order cone $\tilde{\mathcal{K}}_2 = \{ (u,s) \in \mathbb{R}^{m+1} : \|u\|_2\le \kappa s \}$ with $m=3$ and $\kappa = 1/\cos(60^\circ) > 0$:
\begin{equation}
	\Pi_{\tilde{\mathcal{K}}_2}(\tilde u) = 
	\begin{cases}\vspace{-.75mm}
		0 &\text{if\enspace} \|u\|_2\le -  \kappa s\\[1mm]
		\tilde u &\text{if\enspace} \|u\|_2\le  \kappa s\\
		{1\over2}\!\left( 1 + \displaystyle{ \kappa s \over \|u\|_2} \right) [u^\top \!,\,  \|u\|_2/\kappa]^\top
		&\text{if\enspace} \|u\|_2 >  \kappa |s|
	\end{cases}
\end{equation}
for $\tilde u = [u^\top, s]^\top \in \mathbb{R}^{m+1}$ where $u\in\mathbb{R}^{m}$ and $s\in\mathbb{R}$ are a vector of a compatible dimension and a scalar. Notice that a projection $\tilde{u}' = [u^{\prime\top}\!, s']^\top = \Pi_{\tilde{\mathcal{K}}_2}(\tilde u)$ might result in $u^{\prime} \notin\tilde{\mathcal{K}}_2$ if $s' > u_m'$. To handle this, we actually introduce a slack variable $s\in\mathbb{R}$ and consider the constraints in an extended space:
$\tilde{\mathcal{K}} = \tilde{\mathcal{K}}_1 \cap \tilde{\mathcal{K}}_2 \cap \tilde{\mathcal{K}}_3 \subset \mathbb{R}^{m+1}$
where $\tilde{\mathcal{K}}_1 = \{(u,s) \in\mathbb{R}^{m+1} : \| u \|_2 \le 20 \}$ and $\tilde{\mathcal{K}}_3 = \{(u,s)\in\mathbb{R}^{m+1} : e_m^\top u = s\}$.

The cost functions of the corresponding OCP for trajectory generation are defined as
\begin{equation*}
l_f = 500 (\|x_T - [0,4,2]^\top\|_2^2 + \|v_T\|_2^2), \, l_{t} = 0.01\|a_t\|_2^2
\end{equation*}
where $[0,4,2]^\top$ is the target position. The parameters for the MPPI-IPDDP method are given in Tab.~\ref{tab:parameterdrone}.
	
Fig.~\ref{fig:optseqdrone} illustrates the process of generating a smooth trajectory. Fig.~\ref{fig:quadcontrol} compares the noisy control inputs generated by MPPI with the smoothed controls produced by IPDDP. Fig.~\ref{fig:quadcost} demonstrates how the cost and constraint violations decrease over successive iterations of the MPPI-IPDDP method.

\subsection{Comparative Study with Other MPPI Variants}
\label{sec:case:comparison:mppi}
Considering the same scenario of a wheeled mobile robot given in Section~\ref{sec:case:mobile}, we compare the proposed {MPPI-IPDDP} with other existing MPPI methods (vanilla MPPI~\cite{williams2018information}, Log-MPPI~\cite{mohamed2022autonomous} and Smooth-MPPI~\cite{kim2022smooth}) in terms of the computing time and smoothness.

To evaluate smoothness of generated trajectory, the following Mean Squared Curvature (MSC) was used:
\[
{\rm MSC} = \frac{1}{N} \sum_{i=1}^{N-2} k_{i}^{2} \text { where } k_{i}=f^{\prime \prime}\left(x_{i}\right)\,.
\]
At every step of open-loop trajectory generation, we defined the success condition in terms of the computing time $(\leq \tau_{\max})$ and the distance from the target pose $x_{\text {tg}}$, 
$\left\|x_{T}-x_{\text {tg}}\right\|_{2} \leq d_{\epsilon}$
where $\tau_{\max}=10\:{\rm sec}$ and $d_\epsilon=0.1$ are predefined thresholds. 
%

\begin{figure}[t]
	\centering
	\subfloat[Average computing time]{\includegraphics[width=0.488\linewidth]{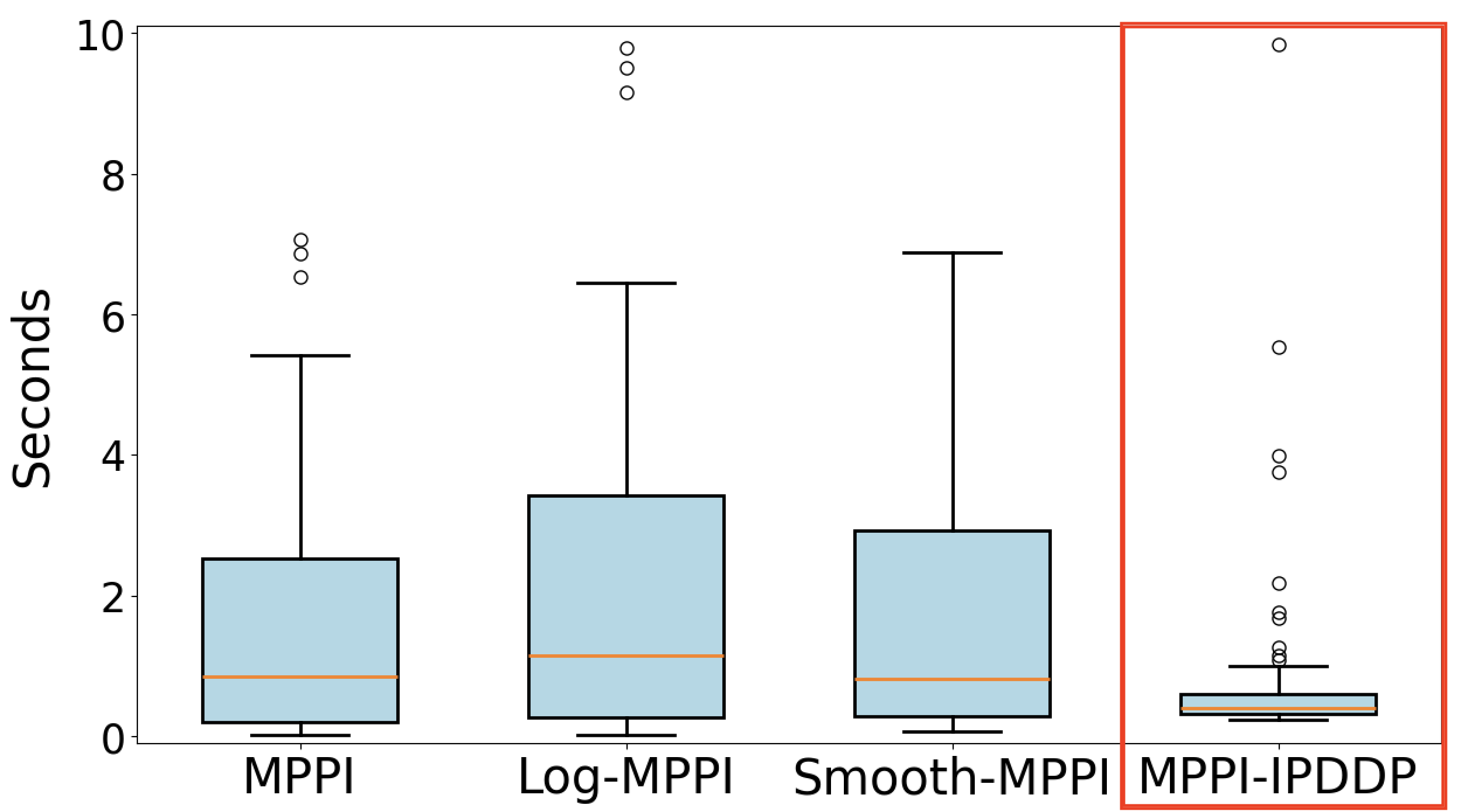}}\hspace{1mm}
	\subfloat[Mean squared curvature\label{fig:8b}]{\includegraphics[width=0.488\linewidth]{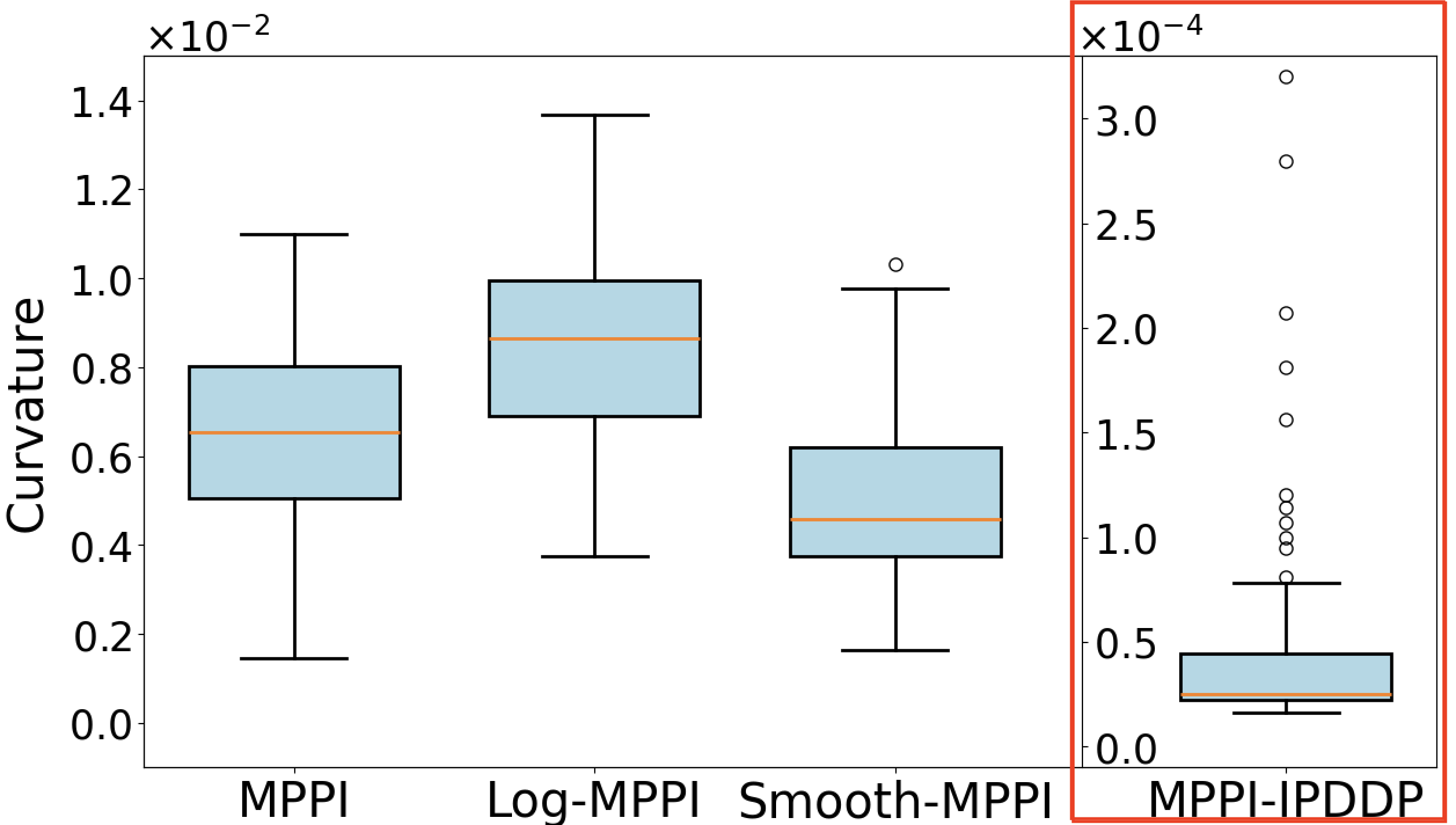}}\vspace{-.75mm}
	\caption{Performance comparisons for MPPI methods.}
	\label{fig:mppiComp}\vspace{-2mm}
\end{figure}

\begin{table}[b]\centering\vspace{-2mm}
	\caption{Comparison of computing time and smoothness for different MPPI methods.}\vspace{-4mm}
	\label{tab:mppiComp:1}
	\renewcommand{\arraystretch}{1.1}
\begin{center}
\begin{tabular}{|@{\!}c@{\!}|c|@{\,\,\,\,}c@{\,\,\,\,}|@{\,\,\,\,}c@{\,\,\,\,}|@{\,\,}c@{\,\,}|@{\,\,}c@{\,\,}|}
\hline
 &  & MPPI & \begin{tabular}{c}
Log- \\
MPPI \\
\end{tabular} & \begin{tabular}{c}
Smooth- \\
MPPI \\
\end{tabular} & \begin{tabular}{c}
MPPI- \\
IPDDP \\
\end{tabular} \\
\hline
\multirow{3}{*}{\begin{tabular}{c}
Avg comp \\
time [sec] \\ 
\end{tabular}} & Q1 & 0.026586 & 0.017963 & 0.073111 & 0.244655 \\
\cline { 2 - 6 }
 & Q2 & 0.855942 & 1.15359 & 0.82165 & 0.408745 \\
\cline { 2 - 6 }
 & Q3 & 5.426678 & 6.451571 & 6.881098 & 1.000769 \\
\hline
\multirow{3}{*}{MSC} & Q1 & 0.001468 & 0.003742 & 0.001631 & 0.000016 \\
\cline { 2 - 6 }
 & Q2 & 0.006526 & 0.008639 & 0.004582 & 0.000025 \\
\cline { 2 - 6 }
 & Q3 & 0.01098 & 0.013666 & 0.009768 & 0.000078 \\
\hline
\end{tabular}
\end{center}
\vspace{-1mm}
{\footnotesize \tiny * Q1, Q2, and Q3 represent the first, second (median), and third quartiles, respectively, of the average computing time and mean squared cost (MSC), calculated from data consisting only of successful simulations. \hfill}
\end{table}

\begin{figure}[t]
\centering
	\subfloat[MPPI]{\includegraphics[width=0.4875\linewidth,height=37mm]{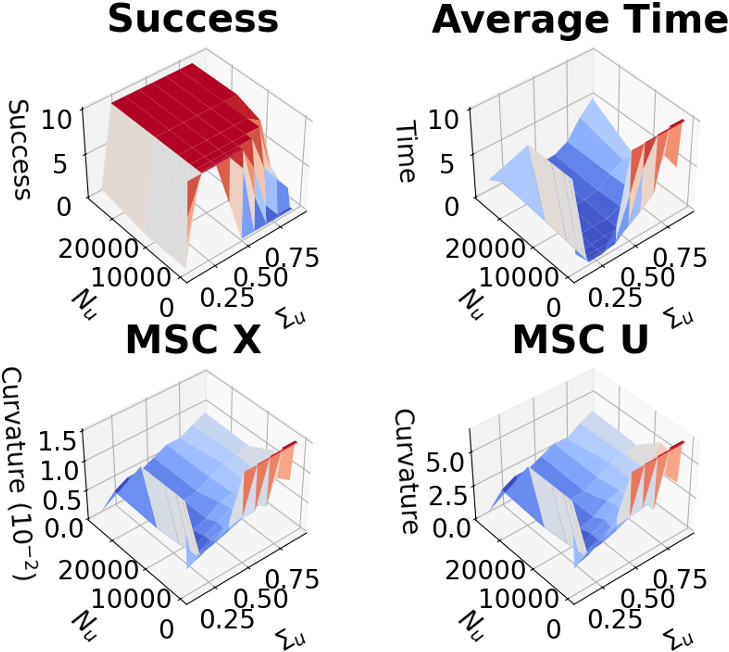}}\hspace{1mm}
	\subfloat[Log-MPPI]{\includegraphics[width=0.4875\linewidth,height=37mm]{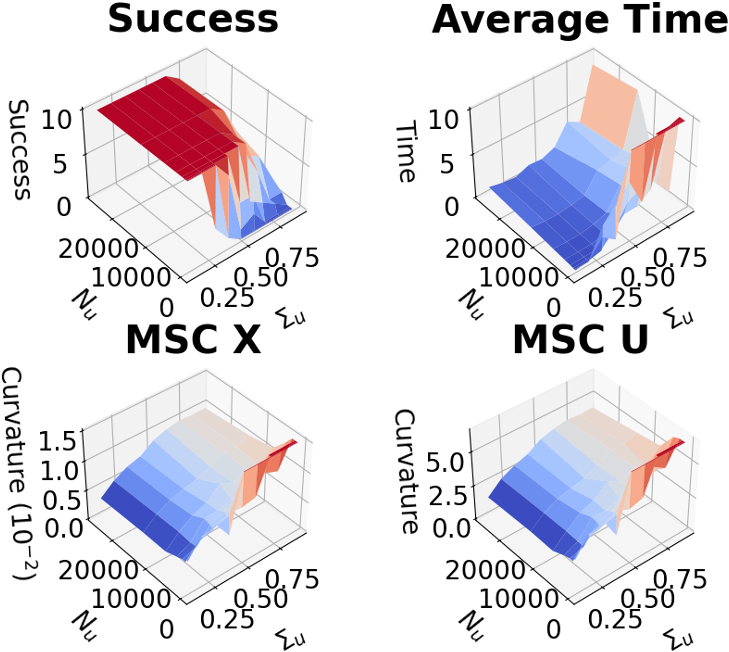}}\vspace{-2mm}
	\subfloat[Smooth-MPPI]{\includegraphics[width=0.4875\linewidth,height=37mm]{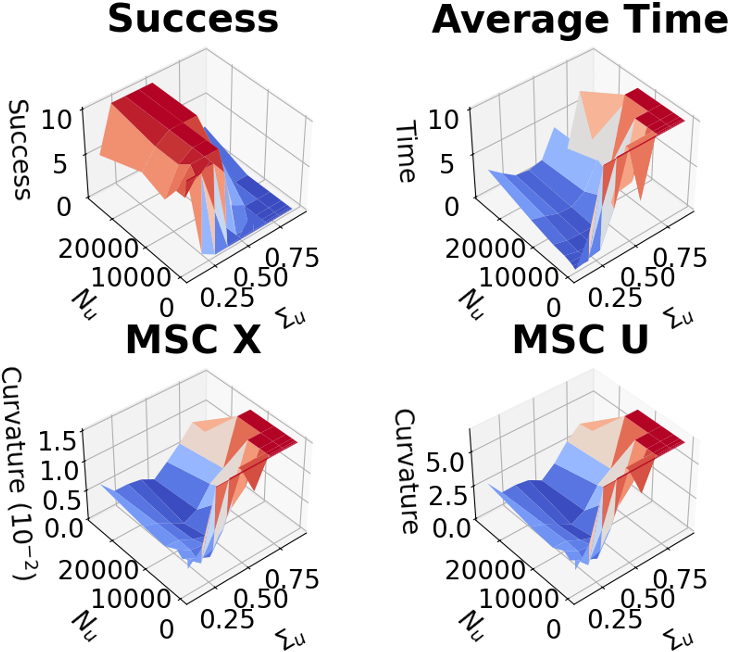}}\hspace{1mm}
	\subfloat[MPPI-IPDDP]{\includegraphics[width=0.4875\linewidth,height=37mm]{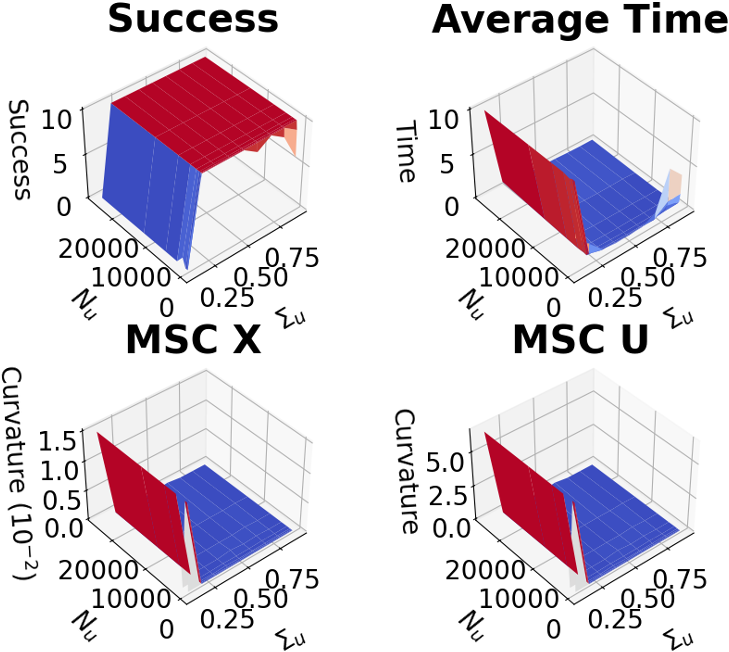}}\vspace{-1mm}
\caption{Performance comparisons of the success rates, average computing time, trajectory MSC for MPPI methods.}
\label{fig:mppiComp:2}\vspace{-2mm}
\end{figure}


\begin{table}[b]\centering\vspace{-2.5mm}
	\caption{Performance comparison of MPPI methods in the BARN dataset~\cite{perille2020benchmarking}.}\vspace{-2mm}
	\label{tab:mppiComp:2}
	\renewcommand{\arraystretch}{1.05}
\begin{tabular}{|@{\,\,}c@{\,\,}|@{\,\,\,\,}c@{\,\,\,\,}|@{\,\,\,\,}c@{\,\,\,\,}|@{\,\,}c@{\,\,}|@{\,\,}c@{\,\,}|}
\hline & MPPI & \begin{tabular}{c} 
Log- \\
MPPI
\end{tabular} & \begin{tabular}{c} 
Smooth- \\
MPPI
\end{tabular} & \begin{tabular}{c} 
MPPI- \\
IPDDP
\end{tabular} \\
\hline $N_u$ & 3200 & 3200 & 12800 & 1600 \\
\hline ${\Sigma}_{u}$ & 0.2 & 0.1 & 0.3 & 0.4 \\
\hline Success ratio [\%] & 97 & 97 & 91.3 & 95.7 \\
\hline Q1 Time [sec] & 0.121522 & 0.169059 & 0.446100 & 0.274369 \\
\hline Q2 Time [sec] & 0.139242 & 0.197344 & 0.511798 & 0.299426 \\
\hline Q3 Time [sec] & 0.179549 & 0.232929 & 0.598198 & 0.346971 \\
\hline MSC & 0.002528 & 0.003089 & 0.005856 & 0.000139 \\
\hline
\end{tabular}
\end{table}

For statistical comparisons of algorithmic performances, numerous simulations with varying parameters of MPPI algorithms were conducted. The number of MPPI samples (\(N_{u}\)) increased from 100 to 25600 by doubling at each step. The covariance matrix of control \(\left(\Sigma_{u}\right)\) varied from \(0.1 I\) to \(0.9 I\), where \(I\) is the identity matrix of a compatible dimension. Fig.~\ref{fig:mppiComp:2} shows the overall performance comparisons of four MPPI methods in terms of the success rate, computing time and trajectory smoothness with different number of samples $N_{u}$ and control covariance $\Sigma_u$. 

Based on the simulation-based statistical analysis presented in Tab.~\ref{tab:mppiComp:1} and Fig.~\ref{fig:mppiComp}, although the first quartile (Q1) statistics of computing time were relatively slow, our MPPI-IPDDP method outperformed the other three MPPI methods in both computing time and the smoothness of trajectory generation. This implies that while the MPPI-IPDDP may have a slower start in some cases, it ultimately provides superior performance overall, achieving faster computations and smoother trajectories compared to the alternative MPPI methods. In addition, the performance of MPPI-IPDDP is less sensitive to changes in the MPPI parameters $N_{u}$ and $\Sigma_u$. This means that the method is more robust and reliable across different settings of these parameters, as illustrated in Fig.~\ref{fig:mppiComp:2}.


We also tested the proposed MPPI-IPDDP in 300 different scenarios of the BARN dataset~\cite{perille2020benchmarking} and compared it with other MPPI methods in terms of success rate, computing time and trajectory MSC. The parameter values of $N_{u}$ and $\Sigma_u$ were customized for each method to optimize its performance. This extensive testing allowed us to assess the robustness and efficiency of the MPPI-IPDDP approach across a wide variety of challenging environments, ensuring that the method was evaluated under diverse conditions. The customized parameters helped each method perform at its best, providing a fair and comprehensive comparison.

The time horizon was set to \({T}=100\), the maximum velocity of \(v_{t}\) was reduced to 1.0 , and each state was defined as \(x_{\text {init }}=[1.5,0, \pi / 2]^{\top}\) and \(x_{\text {tg}}=[1.5,5, \pi / 2]^{\top}\). We expanded the map to be \(5 \mathrm{m}\times3 \mathrm{m}\) from \(3 \mathrm{m}\times3\mathrm{m}\) with additional free space to prevent collision in initial and finish states. The map was also inflated to account for the size of the robot. To properly correspond with the cost calculation \(\mathcal{I}^{\rm PC}(c, r)\) in the Corridor, a distance field was also calculated on the map.

Based on the results of the parameter variation tests, we selected the optimal parameters that yielded the best performance in terms of success rate and smoothness. The results with the BARN dataset indicate that MPPI-IPDDP can generate smooth trajectories in various environments. Although it is more time-consuming than MPPI and Log-MPPI, MPPI-IPDDP produces the smoothest trajectories while using less time compared to Smooth-MPPI.

\subsection{Comparative Study with NLP-based Solvers}
\label{sec:case:comparison:sota}
\begin{figure}[t]
	\centering
	\subfloat[Env. 1: IPOPT fails to generate a collision-free trajectory.]{\includegraphics[width=0.425\linewidth,height=32.5mm]{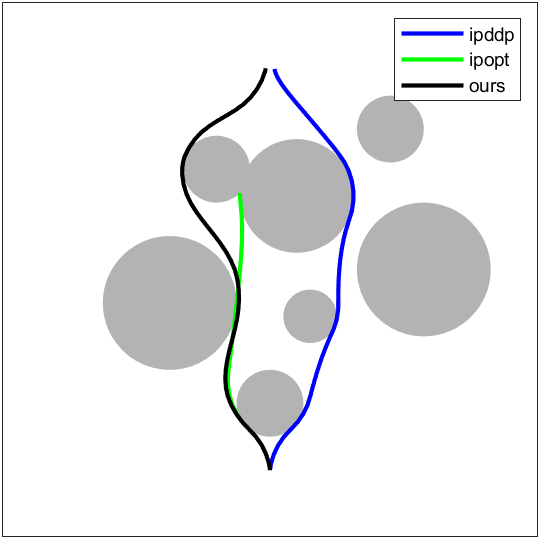}\label{fig:failed}}\hspace{1mm}
	\subfloat[Env. 2: All methods succeed.]{\includegraphics[width=0.425\linewidth,height=32.5mm]{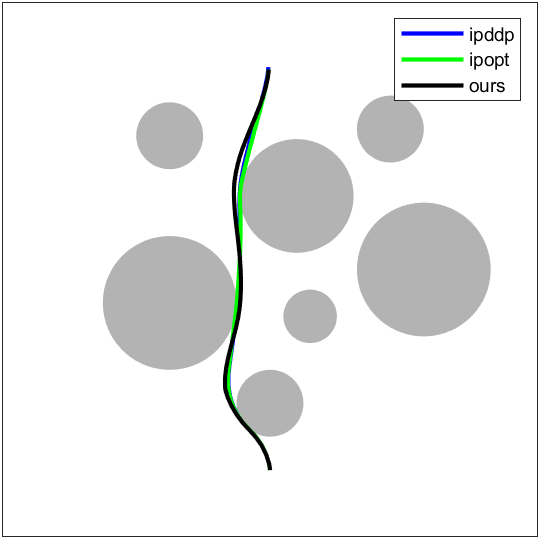}\label{fig:success}}\\[+1.5mm]	
	\subfloat[Open loop control trajectories of Env. 2 in Fig~\ref{fig:success}.]{\includegraphics[width=0.85\linewidth]{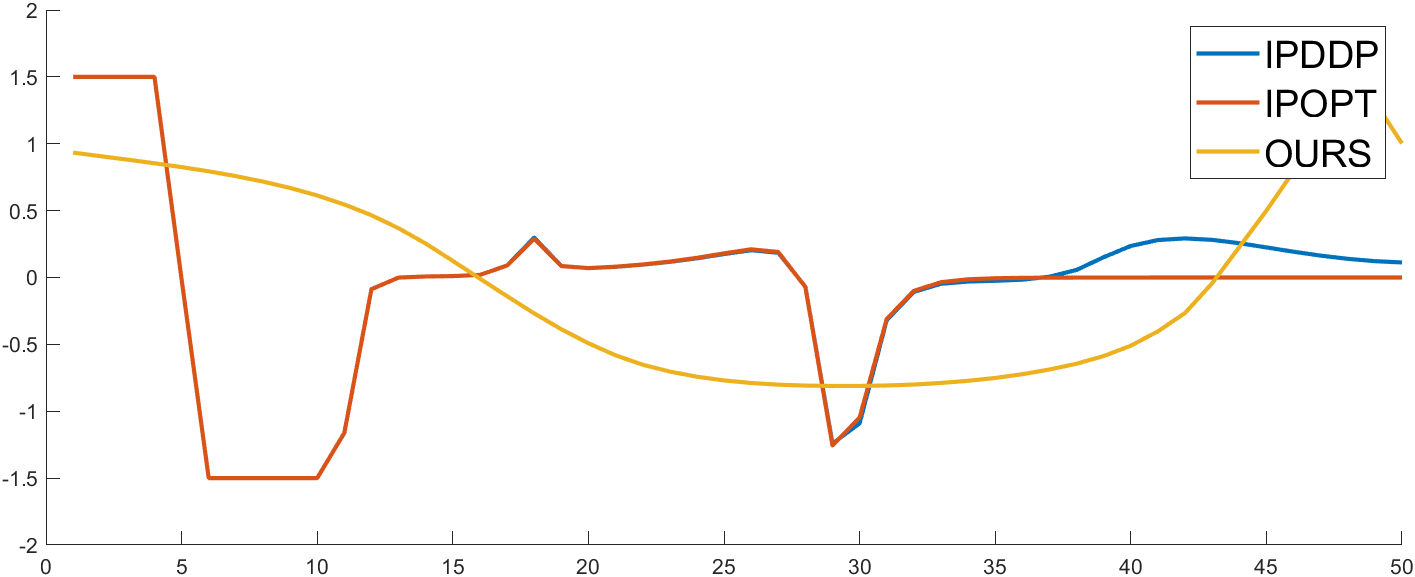}\label{fig:same_input_trajectory}}\\[-.75mm]
	\caption{Comparisons with continuous optimization-based solvers.}
	\label{fig:vs_NLP}\vspace{-2mm}
\end{figure}

\begin{table}[b]\centering\vspace{-2mm}
	\caption{Comparison of computing time and smoothness with NLP-based Solvers.}\vspace{-2mm}
	\label{tab:sotaComp:1}
	\renewcommand{\arraystretch}{1.05}
	\begin{tabular}{|@{\!}c@{\!}|c|@{\,\,\,\,\,\,\,\,}c@{\,\,\,\,\,\,\,\,}|@{\,\,\,\,\,\,\,\,}c@{\,\,\,\,\,\,\,\,}|@{\,\,}c@{\,\,}|}
		\hline
 		&   & IPDDP & IPOPT & MPPI-IPDDP \\
		\hline
		\multirow{3}{*}{\begin{tabular}{c}
		Comp \\
		time [sec] \\ 
		\end{tabular}} 
		& Mean  & 1.64 & 0.117 & 0.539 \\
		\cline { 2 - 5 }
		 & Min  & 1.38 & 0.06 & 0.406 \\
		\cline { 2 - 5 }
		 & Max  & 3.17 & 1.269  & 0.962 \\
		\hline
		\multirow{2}{*}{MSC} & State  &  0.64 & 0.29 & 1.42  \\
		\cline { 2 - 5 }
 		& Control & 0.0530 & 0.0527 & 0.0027  \\
		\hline
	\end{tabular}
\end{table}

In addition to comparisons with other MPPI variants, we also evaluated our hybrid trajectory optimization method against existing state-of-the-art (SOTA) NLP-based methods from a local planning perspective using a receding horizon scheme. Specifically, we compared our method with two baselines: IPOPT~\cite{wachter2006implementation} and IPDDP~\cite{pavlov2021interior}.
For this comparison, we formulated a point-to-point 2D navigation problem for a simple unicycle model in a cluttered environment.\footnote{To ensure a fair comparison, we used MATLAB for all three methods. Specifically, since IPOPT~\cite{wachter2006implementation} and IPDDP~\cite{pavlov2021interior} were implemented using a MATLAB interface, we also employed a MATLAB version of the MPPI-IPDDP algorithm instead of a C++ version. IPOPT is written in C++ and uses a MATLAB interface for problem formulation, while the MPPI-IPDDP used in the comparisons for Tab.~\ref{tab:sotaComp:1} and Fig.~\ref{fig:vs_NLP}~is entirely implemented in MATLAB. Similarly, IPDDP is also written in MATLAB, which leads to slower execution times compared to C++ implementations.}

We treated the obstacle avoidance sub-problem as a constraint for the two gradient-based solvers, considering \(\mathcal{C}^{2}\) smooth ball-type obstacles. We set the same iteration limit and horizon length with random initial guesses for both solvers and our method. Simulations were conducted until the robot reached the desired position in two environments, as shown in Figs.~\ref{fig:failed} and~\ref{fig:success}. Both figures depict closed-loop position trajectories resulting from the implementation of a receding horizon scheme. Due to the dependency of NLP-based solvers on initial guesses, the robot sometimes failed to reach the goal point. Fig.~\ref{fig:failed} illustrates that gradient-based solvers can fail in cases of conflicting gradients, whereas our method can escape these trapped situations regardless of the initial guesses.

To ensure a fair evaluation of computational time and smoothness, we compared the methods in the same environment (Fig.~\ref{fig:success}). Comparisons of the average, minimum, and maximum computing times, as well as the MSC as a smoothness index, are presented in Tab.~\ref{tab:sotaComp:1}. We calculated the MSC for both closed-loop position trajectories and open-loop control input trajectories, particularly for angular velocity. The results show that our method is computationally stable and produces smoother control input trajectories compared to the other methods, as also illustrated in Fig.~\ref{fig:same_input_trajectory}.

\section{Discussion and Future Work}
\label{sec:discussion}

\subsection{Remaining Challenges}
\label{sec:disc:1}
There are still several remaining issues that should be further challenged. 
\paragraph{Real-time implementation}
The proposed algorithm involves three iterative stages, making computation time demanding on a CPU. However, using a GPU for the MPPI stage to leverage massive parallel computation can significantly reduce processing time. The number of iterations needed for IPDDP is relatively low because the initial trajectory input is close to a local optimal solution.

\paragraph{Potential algorithmic failure in dense crowd navigation}
The closer a robot is to obstacles, the higher the likelihood of failure in generating corridors. When a robot makes close contact with obstacles, it becomes challenging to sample a corridor that includes the robot but excludes the obstacle. Alternatively, a soft constraint to keep the robot inside the corridor can be adaptively relaxed by reducing the weight $\lambda_{r}$ in~\eqref{eq:corridor}, whenever the robot gets close to an obstacle.

\paragraph{Planning with uncertainty}
For more precise planning of safety-critical missions, uncertainties induced by modeling errors and external disturbances should be explicitly considered. In our {\tt MPPI-IPDDP} framework, uncertainties could be addressed in the {\tt MPPI}, {\tt Corridor}, or {\tt IPDDP} steps: (a) In {\tt MPPI} with uncertainty, the cost evaluation of~\eqref{eq:mppi} in Alg.~\ref{alg:MPPI} should include a risk-sensitive term that accounts for uncertainties in dynamics and obstacles; (b) In the {\tt Corridor} step with uncertainty, the cost evaluation of~\eqref{eq:corridor} in Alg.~\ref{alg:Corridor} should be modified to account for uncertainties in obstacle configurations; and (c) In {\tt IPDDP} with uncertainty, approaches similar to those used in tube-based robust MPC~\cite{mayne2011tube} and chance-constrained stochastic MPC~\cite{mesbah2016stochastic,kim2013generalised} could be employed to handle uncertainties in planning. However, this may result in conservative constraints due to increasing uncertainty propagation over the horizon.

\paragraph{Planning in dynamic environment}
At the current stage, our focus is on single-robot trajectory optimization, not multi-robot motion planning. In the future, we plan to extend the proposed method to multi-robot trajectory optimization in both cooperative and competitive settings. 
\section{Conclusions}
\label{sec:conclusion}
In this paper, we introduced MPPI-IPDDP, a new hybrid optimization-based local path planning method designed to generate collision-free, smooth, and optimal trajectories. Through two case studies, we demonstrated the effectiveness of the proposed MPPI-IPDDP in environments with complex obstacle layouts. However, there is still room for improvement. As discussed, incorporating Stein Variational Gradient Descent (SVGD) could enhance exploration capabilities. Additionally, addressing planning under uncertainty remains a key challenge. Future work will focus on applying the MPPI-IPDDP algorithm in real-world hardware implementations and integrating it with a global planner.


\bibliographystyle{IEEEtran}
\bibliography{mppi_ddp}

\balance

\end{document}